\documentclass{article}

\usepackage{arxiv}

\usepackage[utf8]{inputenc}
\usepackage[T1]{fontenc}
\usepackage{hyperref}
\usepackage{url}
\usepackage{booktabs}
\usepackage{amsfonts}
\usepackage{nicefrac}
\usepackage{microtype}
\usepackage{relsize}
\usepackage{verbatim}
\usepackage{graphicx}
\usepackage{dcolumn}
\usepackage{bm}
\usepackage{float}
\usepackage{graphics}
\usepackage{color}
\usepackage{xcolor}
\usepackage{soul}
\usepackage{subfigure}
\usepackage{subcaption}
\usepackage{float}
\usepackage{tikz}
\usetikzlibrary{arrows.meta,positioning,fit,calc}
\usepackage{booktabs}
\usepackage{amsmath,amssymb,amsfonts}

\usepackage{algorithm}
\usepackage{algpseudocode}
\usepackage{enumitem}
\setlist[itemize]{leftmargin=*}
\setlist[enumerate]{leftmargin=*}
\definecolor{rev}{rgb}{0,0,0}
\definecolor{rev2}{rgb}{0,0,0}
\usepackage{array}
\newcolumntype{P}[1]{>{\centering\arraybackslash}p{#1}}
\usepackage{multirow}
\usepackage{cancel}
\usepackage[capitalise]{cleveref}

\usepackage{cite}
\usepackage{tabularx}
\usepackage{threeparttable}
\usepackage{pifont}

\newcolumntype{Y}{>{\centering\arraybackslash}X}

\title{The impact of observation density on Bayesian inversion of latent dynamics in shock-dominated flows}

\author{Bipin Tiwari \\
  Department of Mechanical and Aerospace Engineering,\\
  University of Tennessee, Knoxville\\
  Knoxville, TN 37996, USA.\\
  \texttt{btiwari1@vols.utk.edu}\\
  \And
  Muhammad Abid \\
  Department of Mechanical and Aerospace Engineering,\\
  University of Tennessee, Knoxville\\\
  Knoxville, TN 37996, USA.\\
  \texttt{mabid@vols.utk.edu}
  \And
  Omer San \\
  Department of Mechanical and Aerospace Engineering,\\
  University of Tennessee, Knoxville\\\
  Knoxville, TN 37996, USA.\\
  \texttt{osan@utk.edu}
}

\begin{document}
\maketitle
\begin{abstract}
Inferring unknown initial states in shock-dominated compressible flows from sparse measurements is an ill-posed inverse problem, particularly in the presence of limited sensing, measurement noise, and nonlinear wave interactions. In this work, we develop a non-intrusive parametric reduced-order modeling framework for efficient Bayesian initial-state inversion with uncertainty quantification. The proposed approach combines a convolutional autoencoder with a learned latent-space forward operator. The autoencoder compresses high-dimensional flow fields into a compact nonlinear latent representation, while the forward operator maps encoded initial conditions to final-time latent states. This AE-ROM surrogate enables rapid forward prediction and can be embedded directly within a No-U-Turn Sampler (NUTS) for posterior exploration. The framework is demonstrated using 500 high-fidelity Sod shock tube simulations generated through Latin hypercube sampling and solved with a fifth-order WENO scheme. The inverse problem seeks to recover the unknown left and right initial density and pressure states from sparse, noisy observations of the final-time density and pressure fields. Results show that the AE-ROM accurately reconstructs the dominant shock-tube structures, including the rarefaction wave, contact discontinuity, and shock front. A latent dimension of 32 provides a practical balance between reconstruction accuracy and reduced-space compactness, while a training budget of 250 simulations is sufficient to achieve high-fidelity reconstructions. Increasing the observation density produces substantial posterior uncertainty contraction, with the mean posterior standard deviation decreasing by approximately 78\% for density and 76\% for pressure, while the posterior mean error decreases more moderately. These findings indicate that denser observations primarily improve parameter identifiability and confidence in the inferred state. Overall, the proposed AE-ROM-enabled Bayesian framework provides a computationally tractable and uncertainty-aware approach for inverse analysis of shock-dominated flows, establishing a foundation for future extensions to multidimensional compressible flows and high-speed aerodynamic digital-twin applications.
\end{abstract}


\section{\label{sec:level1}Introduction}

High-speed compressible flows are the fundamental aspect of modern aerospace engineering, underpinning the design and analysis of hypersonic vehicles, atmospheric reentry capsules, and advanced propulsion systems \cite{Anderson2017, Bertin2021, Heiser2018, Han2024}. These flow regimes are fundamentally defined by the presence of strong nonlinear phenomena, including shock waves, contact discontinuities, expansion fans, and intricate shock-boundary layer interactions \cite{Toro2009, LeVeque2002}. The physics governing these flows is exceptionally complex, exhibiting multi-scale dynamics and a profound sensitivity to initial and boundary conditions \cite{Fedorov2019, Wang2018}. Even small perturbations can lead to significant, often non-intuitive, changes in the flow structure, such as shock oscillations or transitions in aerodynamic stability \cite{Dolling2020, Knight2018}. This inherent complexity poses immense challenges for both forward prediction and system analysis.

A particularly challenging and vital task within this domain is the solution of inverse problems \cite{Tarantola2005}. In contrast to forward problems where system outputs are computed from known inputs, inverse problems seek to infer unknown causal factors, such as initial conditions, boundary parameters, or geometric properties, from a set of observed system responses \cite{Kaipio2006, Aster2018}. In the context of high-speed flows, this could involve determining the freestream conditions that produced an observed shock structure, reconstructing an entire flow field from sparse sensor measurements, or estimating the aerodynamic coefficients of a vehicle based on its trajectory data \cite{deFaria2023, Pan2020}.

The ability to solve such inverse problems efficiently is critical for numerous applications. It is essential for validating and calibrating numerical models against experimental data, a process known as model updating or data assimilation \cite{Reich2015}. Furthermore, it forms the basis of system identification, design optimization, and health monitoring \cite{Ghattas2021, Allaire2019}. As the aerospace industry moves towards the paradigm of the digital twin, a high-fidelity, living model of a physical asset, the capacity to solve inverse problems in near real time to update the twin with operational data becomes a key enabling technology \cite{Kapteyn2021, Rasheed2020}.

However, solving inverse problems for high-speed flows remains fraught with difficulty, primarily due to the prohibitive computational cost of high-fidelity simulations. Numerical methods such as Direct Numerical Simulation (DNS), Large Eddy Simulation (LES), and Reynolds-Averaged Navier-Stokes (RANS) simulations with advanced schemes like Weighted Essentially Non-Oscillatory (WENO) are required to accurately capture the sharp gradients and discontinuities inherent in these flows \cite{Shu2019, Pirozzoli2021}. These simulations demand extremely fine computational grids and small time steps, leading to computational runtimes on the order of hours, days, or even weeks on high-performance computing clusters \cite{Choi2021}.

This computational expense creates an intractable bottleneck for inverse problems, which are fundamentally multi-query tasks. Whether employing gradient-based optimization methods or probabilistic sampling techniques, solving an inverse problem necessitates thousands, or even millions, of forward model evaluations to adequately explore the high-dimensional parameter space \cite{Smith2013}. Performing this exploration with full-order computational fluid dynamics (CFD) solvers is, in most practical scenarios, computationally infeasible.

To overcome this computational challenge, the field has increasingly turned towards the Reduced-Order Models (ROMs) \cite{Benner2015, Rowley2017}. The central premise of a ROM is to approximate the dynamics of a high-dimensional system within a low-dimensional subspace, thereby creating a computationally inexpensive surrogate that can be evaluated rapidly \cite{Quarteroni2016}. Traditional projection-based ROMs, most notably those based on Proper Orthogonal Decomposition (POD) coupled with Galerkin projection, have achieved considerable success \cite{tang2025applications, dai2025joint}. These methods construct a set of optimal, data-derived basis functions (POD modes) and project the governing equations onto the subspace spanned by these modes.

Despite their strengths, POD-Galerkin ROMs face limitations when applied to strongly nonlinear, advection-dominated systems like shock-driven flows. The linear superposition principle underlying POD struggles to efficiently represent the transport of localized, sharp features, often requiring a large number of modes and suffering from stability issues \cite{Carlberg2017, Ohlberger2016}. Furthermore, the Galerkin projection step is intrusive, requiring modification of the source code of the full-order solver, which can be a complex and nontrivial process.

The recent integration of machine learning and scientific computing has driven the development of non-intrusive, data-driven ROMs that can overcome these limitations \cite{Brunton2020, Duraisamy2019}. These methods learn the low-dimensional dynamics directly from snapshots of high-fidelity simulation data. Recent advances in spectral embedding plus multi-resolution neural operator learning have shown pretty strong capabilities for nonlinear reduced-order modeling and for scientific machine learning of complex flow systems  \cite{abid2026simr,abid2025spectral}. Among these techniques, deep learning models, particularly autoencoders, have emerged as an exceptionally powerful tool for nonlinear dimensionality reduction \cite{Goodfellow2016, LeCun2015}. An autoencoder consists of two neural networks: an encoder that compresses the high-dimensional input data into a low-dimensional latent space, and a decoder that reconstructs the original data from this compressed representation. By training on flow-field data, an autoencoder can learn a compact latent representation that efficiently captures the essential nonlinear features of the system, such as the position and strength of shock waves \cite{Maulik2021, Mohan2018}.

While an accurate and fast ROM addresses the challenge of forward model evaluation, another critical aspect of real-world inverse problems is the quantification of the uncertainty. Experimental measurements are invariably sparse and corrupted by noise, and the models themselves are imperfect representations of reality \cite{Sullivan2015}. Consequently, a deterministic point estimate of the unknown parameters is often insufficient. A robust solution requires a full characterization of the uncertainty associated with the inferred parameters, which is crucial for reliable decision-making and risk assessment.

The Bayesian inference paradigm provides a rigorous mathematical framework for solving inverse problems under uncertainty \cite{Gelman2013}. Bayes' theorem elegantly combines prior knowledge about the parameters with the likelihood of the observed data (given a set of parameters) to yield the posterior probability distribution. This posterior distribution encapsulates our complete updated knowledge of the parameters, inherently providing a full quantification of their uncertainty \cite{Sivia2006}.

For complex, nonlinear physical models, the posterior distribution is typically high-dimensional and cannot be determined analytically. Its characterization relies on computational techniques, primarily Markov Chain Monte Carlo (MCMC) methods \cite{Brooks2011}. MCMC algorithms construct a Markov chain whose stationary distribution is the target posterior, allowing one to draw samples and approximate the distribution. Modern MCMC algorithms, such as Hamiltonian Monte Carlo (HMC) and its adaptive variant, the No-U-Turn Sampler (NUTS), are particularly effective at exploring complex, high-dimensional parameter spaces efficiently \cite{Hoffman2014, Betancourt2017}.

This brings us to the synergistic core of the proposed work: the integration of machine learning-based ROMs with advanced Bayesian inference. The prohibitive cost of CFD makes direct MCMC sampling, where each step in the chain would require a full-order simulation impossible. By replacing the expensive high-fidelity forward model with a fast and accurate ROM surrogate inside the MCMC loop, the Bayesian inverse problem is rendered computationally tractable \cite{Cui2016, Marzouk2022}. This powerful combination allows for both rapid parameter inference and rigorous uncertainty quantification.

In this paper, we develop and demonstrate a parametric ROM framework designed specifically for solving inverse problems in high-speed, shock-dominated flows. We use the canonical one-dimensional Sod shock tube problem as a challenging test case that embodies the essential physics of interest, including a shock wave, a contact discontinuity, and a rarefaction wave. Our methodology involves three primary stages. First, we generate a comprehensive high-fidelity dataset by simulating the shock tube problem for a wide range of initial conditions sampled via Latin Hypercube Sampling (LHS). Second, we develop a convolutional autoencoder-based ROM (AE-ROM). This model not only learns to compress and reconstruct the flow-field snapshots but also incorporates a forward operator a multilayer perceptron, that learns the direct mapping from the initial conditions to the final state within the low-dimensional latent space. This creates a fully non-intrusive surrogate capable of rapid forward predictions.

Finally, we integrate this AE-ROM surrogate into a Bayesian inference framework. Using the No-U-Turn Sampler, we solve the inverse problem of inferring the unknown initial pressure and density conditions from sparse, noisy observations of the final flow state. We demonstrate that our framework can accurately recover the posterior distributions of the unknown parameters, providing both precise mean estimates and credible intervals that quantify the estimation uncertainty. The significant computational speedup afforded by the ROM makes this probabilistic approach practical and efficient.

The remainder of this paper is organized as follows. Section 2 details the governing Euler equations for compressible flow and the numerical setup used to generate the high-fidelity data. Section 3 presents our methodology in depth, covering the architecture of the autoencoder and forward operator, the formulation of the Bayesian inverse problem, and the implementation of the MCMC sampler. Section 4 presents and discusses the results, focusing on the reconstruction accuracy of the AE-ROM, the performance of the inverse inference, and a parametric analysis of the framework. Finally, Section 5 provides concluding remarks and outlines promising directions for future research, including extensions to multi-dimensional problems and applications in reentry vehicle aerodynamics.

\section{Motivation and Background}
\begin{figure}[ht!]
    \centering
    \includegraphics[width=0.65\textwidth]{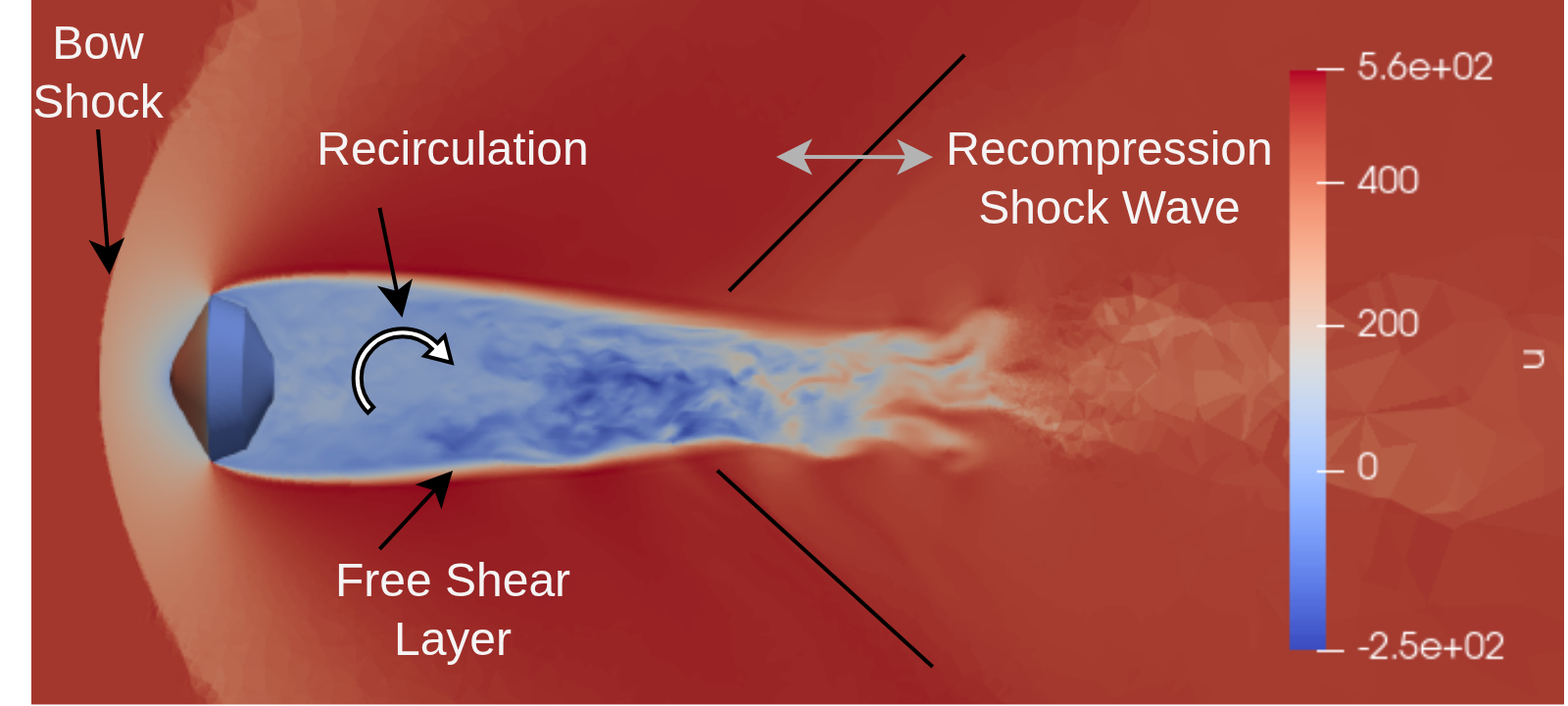}
    \caption{Mach 1.44 flow field around a Genesis sample return capsule model, showing the major shock-dominated flow structures. The labeled features include the detached bow shock, separated recirculation region, free shear layer, and recompression shock wave in the wake. Adapted from Kassem et al.~\cite{kassem2026discovery}.}
    \label{fig:hayabusa}
\end{figure}

\begin{figure}[ht!]
    \centering
    \includegraphics[width=0.65\textwidth]{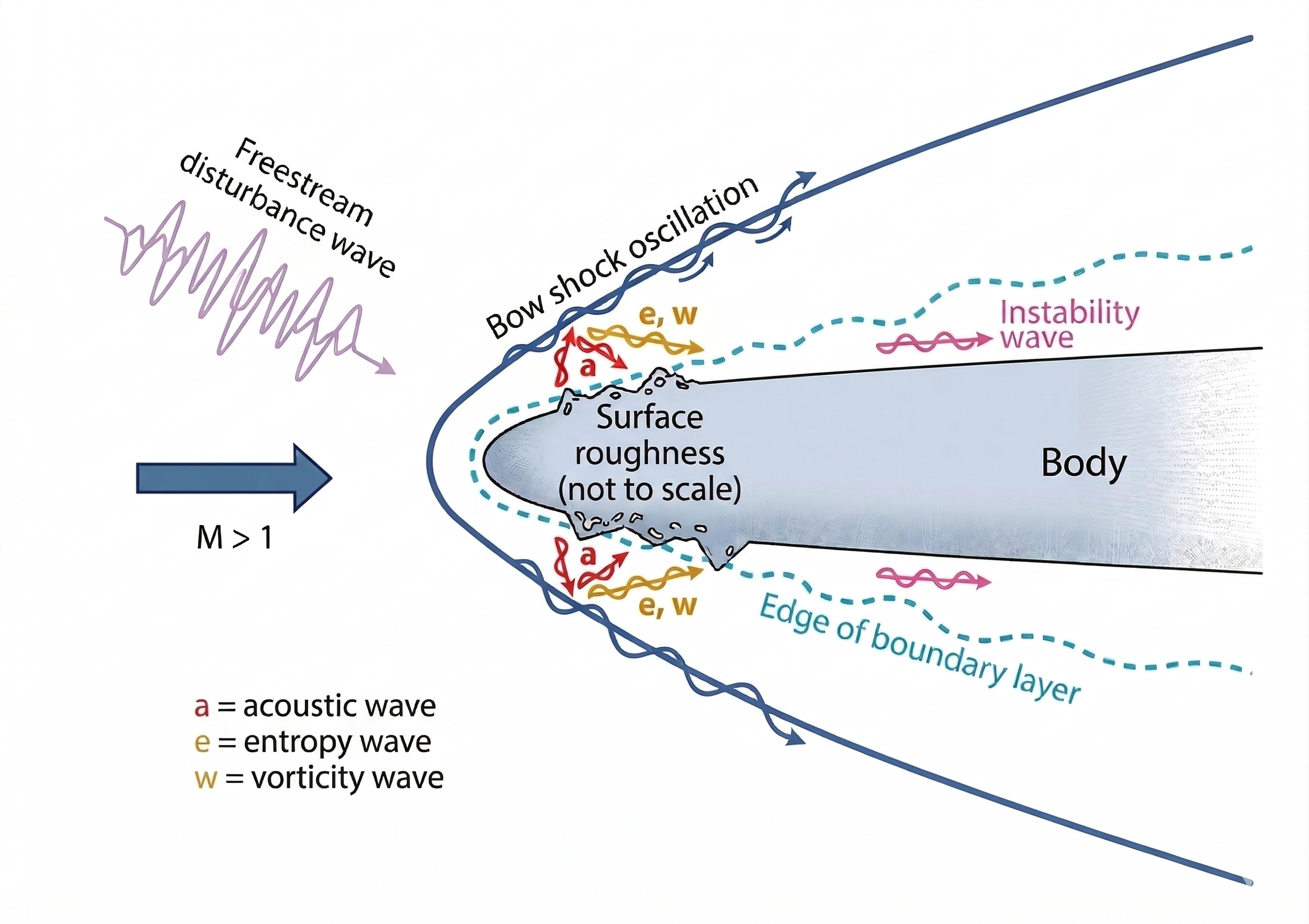}
    \caption{Schematic of a blunt body exposed to freestream disturbances in supersonic flow. Incoming disturbances interact with the bow shock and surface roughness, generating acoustic, entropy, and vorticity waves that can amplify within the boundary layer and contribute to downstream instability waves. Adapted from Wang et al.~\cite{Wang2018}.}
    \label{fig:disturbance}
\end{figure}

High-speed flows are defined by their complex and highly nonlinear nature. The governing physics, which includes phenomena such as shock waves, discontinuities, and rarefaction waves, introduces strong nonlinearities that are challenging to model and predict. As illustrated by the intricate flow field around a reentry capsule in Figure~\ref{fig:hayabusa}, the interaction of bow shocks, recirculation region, shear layers, recompression shock wave, and vortex shedding creates a dynamically rich environment. Furthermore, these systems exhibit a high degree of sensitivity; small changes in freestream conditions or boundary parameters can drastically alter the flow field, leading to different shock structures and stability characteristics, as depicted schematically in Figure~\ref{fig:disturbance}. This strong dependence on small changes makes direct measurements difficult and makes it especially hard to determine the underlying parameters from limited data, which is the core challenge of the inverse problem.

\subsection{Challenges In Inverse Problems}

Solving inverse problems in high-speed aerodynamics is hindered by two primary challenges: data scarcity and computational cost. Experimental facilities like wind tunnels and flight tests can provide invaluable data, but measurements are often sparse, restricted to a few sensor locations, and inevitably corrupted by noise. It is difficult to directly observe and quantify critical flow interactions across the entire domain, leaving significant portions of the flow field uncharacterized.

\begin{figure}[ht!]
    \centering
    \includegraphics[width=0.65\textwidth]{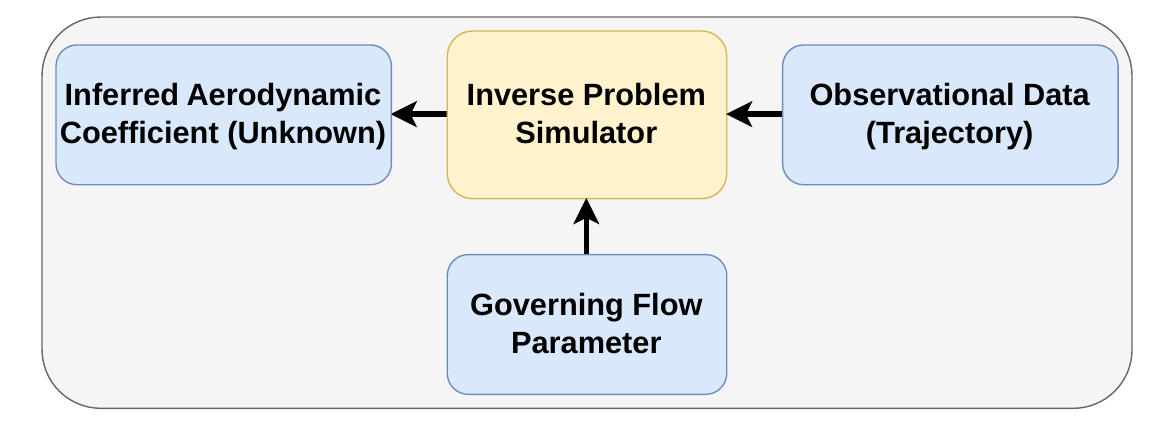}
    \caption{Conceptual diagram of an inverse problem framework. The simulator (forward model) is queried repeatedly with candidate unknown parameters (e.g., aerodynamic coefficients) to generate flow predictions, which are then compared against trajectory or sensor data.}
    \label{fig:inverse_simulator}
\end{figure}

To fill these observational gaps, one must rely on numerical simulations. However, as previously discussed, high-fidelity CFD solvers are computationally expensive. This cost becomes prohibitive when tackling inverse problems, which are iterative by nature. As conceptualized in Figure~\ref{fig:inverse_simulator}, an inverse problem solver repeatedly queries a forward model (the "simulator"). It proposes a set of unknown parameters (e.g., aerodynamic coefficients), runs the forward simulation to obtain the corresponding flow parameters, and compares this output to the observed data. This process is repeated thousands of times to find the parameters that best explain the observations. When the forward model is a full-order CFD simulation, this iterative loop is computationally intractable for all but the simplest cases. This severe computational bottleneck motivates a fundamental shift away from direct simulation.

\subsection{Reduced-Order Modeling Approach}

Reduced-Order Models (ROMs) offer a powerful solution to this computational challenge. Instead of solving the full system of partial differential equations, ROMs approximate the flow dynamics using a low-dimensional model that captures the most dominant features of the flow. By operating in a highly compressed state space, these models can run orders of magnitude faster than the full-order CFD solver they are derived from. This dramatic speedup transforms the inverse problem from an intractable task to a feasible one.

The integration of ROMs into an inverse problem framework enables the rapid, repeated forward evaluations necessary for parameter estimation and optimization. By leveraging a fast and accurate ROM surrogate, it becomes possible to perform thousands of forward solves in minutes or hours, rather than weeks or months. This efficiency not only accelerates the solution process but also enables more sophisticated analyses, such as comprehensive uncertainty quantification using MCMC methods, which would be completely out of reach with traditional solvers. The development of a robust, parametric ROM is therefore the key enabling step for building an efficient framework to solve inverse problems in high-speed flows.

\section{Methodology}
The proposed framework integrates high-fidelity data generation, a non-intrusive deep learning-based ROM, and a Bayesian inference scheme to solve inverse problems for high-speed flows. This section details each component of our methodology, from the governing equations of the test problem to the formulation of the probabilistic inference.

\subsection{Canonical Test Problem: Sod Shock Tube}
To develop and validate our framework, we utilize the Sod shock tube problem, a canonical one-dimensional Riemann problem that serves as a standard benchmark for compressible flow solvers and nonlinear model analysis \cite{Toro2009}. The problem is governed by the 1D Euler equations, which express the conservation of mass, momentum, and energy for an inviscid, non-heat-conducting fluid. In their conservative form, the equations are written as:
\begin{equation}
\frac{\partial \mathbf{U}}{\partial t} + \frac{\partial \mathbf{F}(\mathbf{U})}{\partial x} = 0,
\end{equation}
where $\mathbf{U}$ is the vector of conserved state variables and $\mathbf{F}(\mathbf{U})$ is the flux vector:
\begin{equation}
\mathbf{U} = \begin{bmatrix} \rho \\ \rho u \\ E \end{bmatrix}, \quad
\mathbf{F}(\mathbf{U}) = \begin{bmatrix} \rho u \\ \rho u^2 + p \\ u(E+p) \end{bmatrix}.
\end{equation}
Here, $\rho$ is the fluid density, $u$ is the velocity, $p$ is the pressure, and $E$ is the total energy per unit volume, defined for an ideal gas as:
\begin{equation}
E = \frac{p}{\gamma - 1} + \frac{1}{2}\rho u^2,
\end{equation}
with $\gamma$ being the ratio of specific heats, taken as 1.4 for air.

\begin{figure}[ht!]
    \centering
    \includegraphics[width=0.75\textwidth]{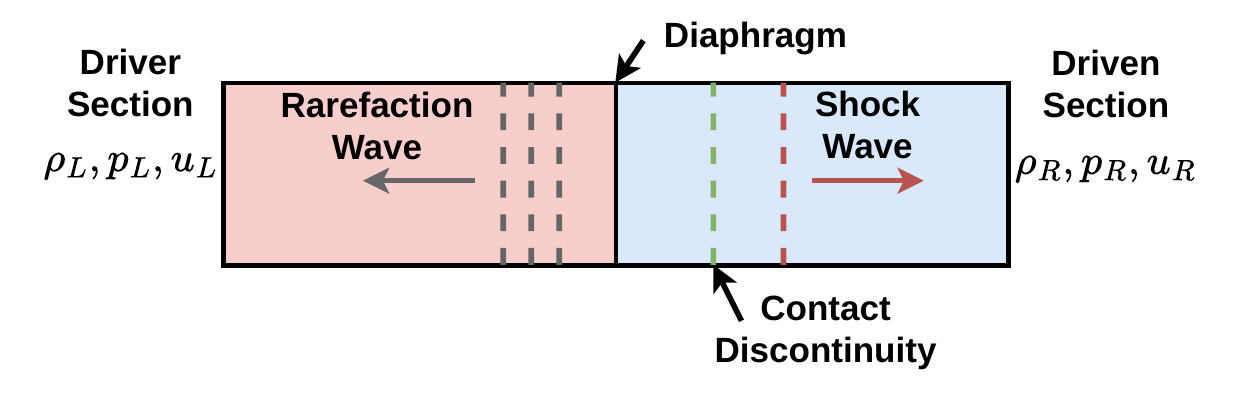}
    \caption{Schematic of the Sod shock tube problem after the diaphragm is removed, showing the resulting rarefaction wave, contact discontinuity, and shock wave.}
    \label{fig:sod_schematic}
\end{figure}

The initial condition consists of two distinct, uniform states of the fluid separated by a diaphragm at $x=0.5$. When the diaphragm is instantaneously removed at $t=0$, the system evolves into a characteristic wave pattern consisting of three main features: a right-traveling shock wave, a contact discontinuity, and a left-traveling rarefaction (or expansion) wave, as illustrated in Figure~\ref{fig:sod_schematic}. This setup provides a robust testbed for evaluating a model's ability to capture sharp discontinuities and smooth expansion profiles simultaneously.

\subsection{Data Generation}

To train and validate the proposed reduced-order modeling framework, we constructed a comprehensive dataset of high-fidelity solutions to the one-dimensional Sod shock tube problem under varying initial conditions. The objective of this dataset is to span a broad parametric space of physically meaningful Riemann problems while maintaining controlled variability for inverse inference.

The parameter space was sampled using latin hypercube sampling (LHS), a stratified sampling technique that ensures uniform coverage of multidimensional domains. The sampled parameters include the left-state density ($\rho_L$), left-state pressure ($p_L$), right-state density ($\rho_R$), and right-state pressure ($p_R$). The initial velocity was set to zero throughout the domain for all cases ($u_L = u_R = 0$). The sampling ranges were selected to generate physically admissible shock-tube configurations while allowing variations in shock strength and wave interactions.

\begin{table}[h!]
\centering
\caption{Parameter ranges used for LHS of initial conditions.}
\label{tab:sampling}
\renewcommand{\arraystretch}{1.2}
\begin{tabular}{|c|c|c|}
\hline
\textbf{Parameter} & \textbf{Left State ($x \le 0.5$)} & \textbf{Right State ($x > 0.5$)} \\ 
\hline
Density ($\rho$)  & $0.5 \le \rho_L \le 1.5$  & $0.05 \le \rho_R \le 0.15$ \\ 
\hline
Velocity ($u$)    & $u_L = 0$                 & $u_R = 0$ \\ 
\hline
Pressure ($p$)    & $0.5 \le p_L \le 1.5$     & $0.05 \le p_R \le 0.15$ \\ 
\hline
\end{tabular}
\end{table}

\begin{figure*}[ht!]
    \centering
    \includegraphics[width=0.95\textwidth]{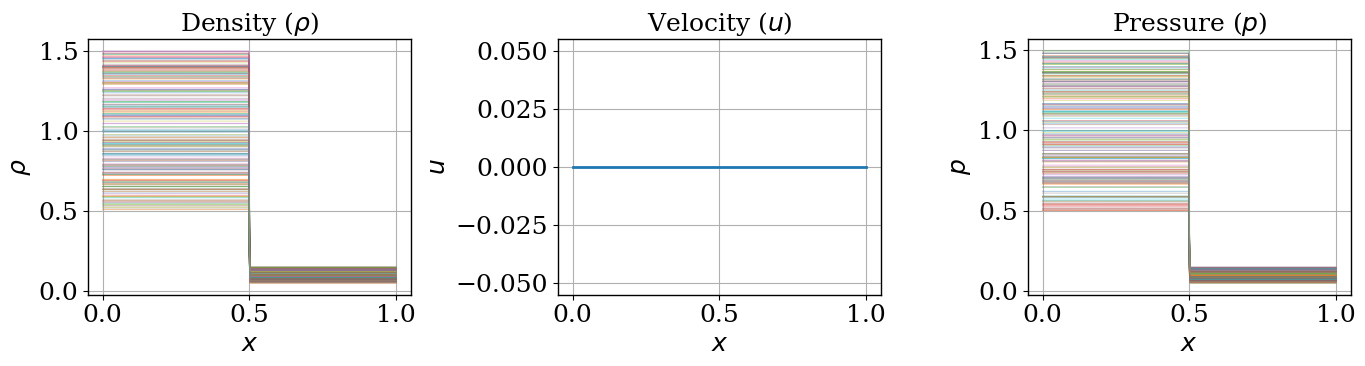}
    \caption{Ensemble of initial conditions generated by Latin hypercube sampling for the Sod shock tube simulations. The density and pressure fields vary across the left and right states, while the velocity is initialized to zero throughout the domain for all cases. Each curve represents one sampled Riemann problem with a discontinuity at the diaphragm location.}
    \label{fig:initial_ensemble}
\end{figure*}

A total of 500 distinct initial conditions were generated using LHS. Each initial condition defines a unique Riemann problem characterized by a discontinuity at the diaphragm location ($x = 0.5$) separating two constant thermodynamic states. Figure~\ref{fig:initial_ensemble} illustrates the ensemble of sampled initial conditions for density, velocity, and pressure. Each colored curve corresponds to one simulation instance. The density and pressure fields exhibit substantial variability across the left and right states, while velocity remains identically zero throughout the domain. This diversity ensures that the dataset captures a wide range of shock intensities and thermodynamic contrasts, which is essential for learning a robust parametric reduced manifold.

For each sampled initial condition, the compressible Euler equations were solved using a high-fidelity finite-volume solver. The spatial domain $[0,1]$ was discretized using 1000 uniform grid points. A fifth-order Weighted Essentially Non-Oscillatory (WENO) scheme was employed for spatial reconstruction, coupled with an HLLC Riemann solver for numerical flux evaluation. Time integration was performed using a third-order Runge--Kutta scheme. Simulations were advanced until a final time of $t = 0.2\,\text{s}$, by which time the shock, contact discontinuity, and rarefaction wave were fully developed and well separated.

\begin{table}[h!]
\centering
\caption{High-fidelity numerical solver configuration.}
\label{tab:solver}
\renewcommand{\arraystretch}{1.2}
\begin{tabular}{|l|c|}
\hline
\textbf{Parameter} & \textbf{Setting} \\ 
\hline
Numerical Scheme & 5th-Order WENO \\ 
\hline
Flux Solver & HLLC Riemann Solver \\ 
\hline
Time Integration & 3rd-Order Runge--Kutta (RK3) \\ 
\hline
Grid Points ($n_x$) & 1000 \\ 
\hline
Domain ($x$) & $[0,1]$ \\ 
\hline
Final Time ($t_f$) & $0.2$ s \\ 
\hline
\end{tabular}
\end{table}

The resulting dataset consists of full-field snapshots of density, velocity, and pressure at the final time for each simulation. These high-dimensional flow states serve as training targets for the autoencoder component of the AE-ROM. Additionally, both initial and final states are encoded into the latent space to train the forward operator that maps initial conditions to final latent representations.

This dataset construction ensures (i) sufficient variability to learn a parametric reduced manifold, (ii) preservation of sharp hyperbolic features such as shocks and contact discontinuities, and (iii) controlled coverage of the inverse parameter space to enable stable Bayesian inference.

\subsection{Autoencoder-Based Reduced-Order Model}

\begin{figure}[ht!]
    \centering
    \includegraphics[width=0.65\textwidth]{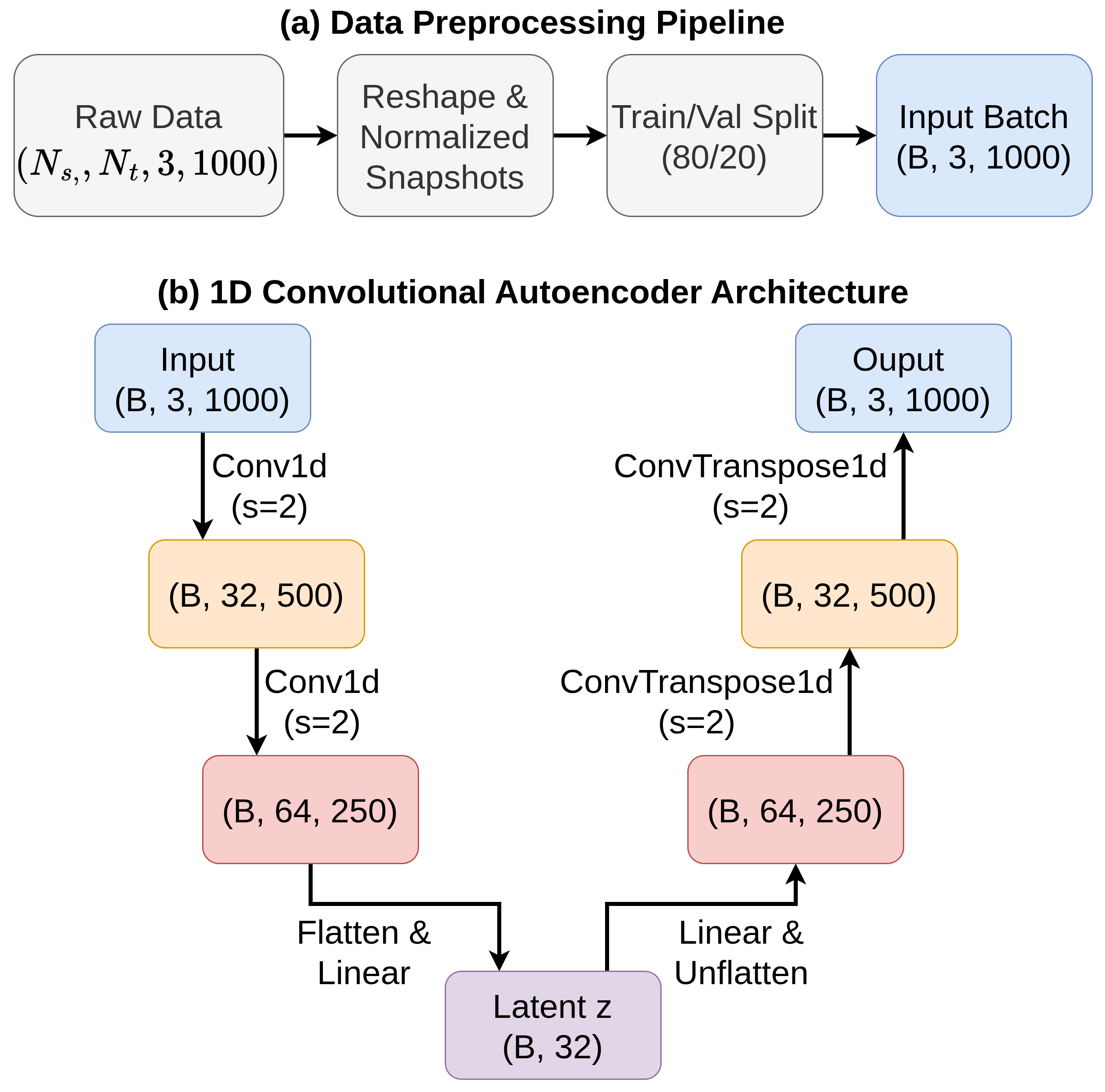}
    \caption{Data preprocessing and one-dimensional convolutional autoencoder architecture used for reduced-order representation of the shock-tube flow fields. The raw simulation data are reshaped, normalized, and divided into training and validation sets before being passed to the network in mini-batches. The encoder compresses each density, velocity, and pressure snapshot into a 32-dimensional latent vector, while the decoder reconstructs the full field with the original spatial resolution.}
    \label{fig:autoencoder}
\end{figure}

\begin{figure}[ht!]
    \centering
    \includegraphics[width=0.65\textwidth]{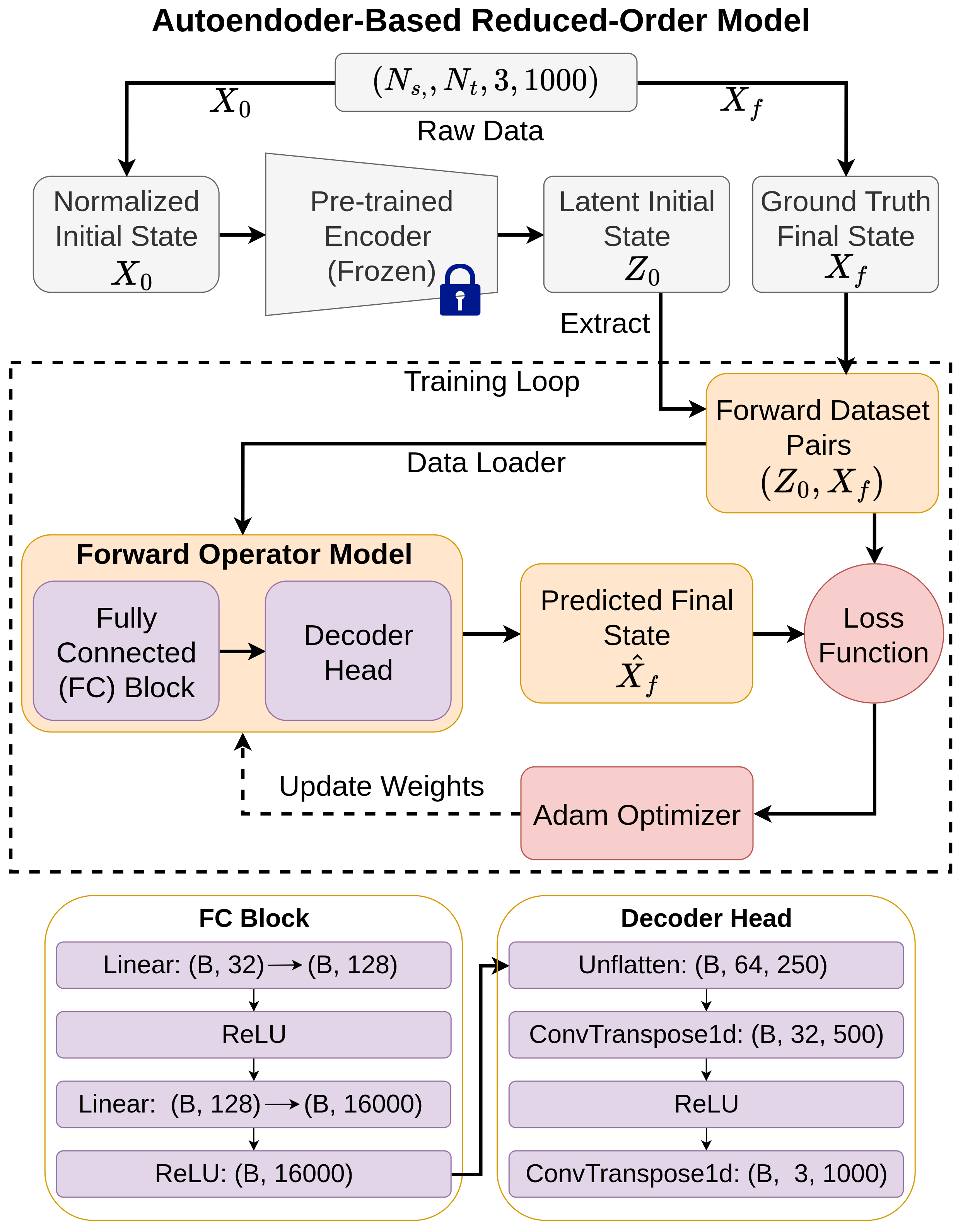}
    \caption{Training workflow for the AE-ROM forward operator. The pretrained encoder is frozen and used to map normalized initial states into latent initial representations. These latent states are paired with corresponding high-fidelity final states to train a forward operator consisting of a fully connected block and decoder head. The predicted final state is compared with the ground-truth final state through a reconstruction loss, and the model parameters are updated using the Adam optimizer.}
    \label{fig:ae_rom_workflow}
\end{figure}

The reduced-order modeling framework combines nonlinear dimensionality reduction with latent-space evolution to construct a computationally efficient surrogate for shock-dominated compressible flows. The convolutional autoencoder architecture and the overall reduced-order modeling workflow are illustrated in Figures~\ref{fig:autoencoder} and~\ref{fig:ae_rom_workflow}.

The high-fidelity dataset consists of numerical solutions of the Sod shock tube problem. Each flow snapshot contains three physical variables, namely density ($\rho$), velocity ($u$), and pressure ($p$), defined over 1000 spatial grid points along the one-dimensional computational domain. Accordingly, each flow state is represented as a tensor of size $(3,1000)$, where the first dimension corresponds to the three physical variables and the second dimension corresponds to the spatial discretization. During neural network training, mini-batches of size $B$ are constructed, resulting in input tensors of shape $(B,3,1000)$.

Nonlinear dimensionality reduction is performed using the one-dimensional convolutional autoencoder shown in Figure~\ref{fig:autoencoder}. Let $X$ denote a high-dimensional flow state of size $(3,1000)$. The encoder, denoted by $\Phi$, maps this state to a compact latent representation defined as
$z = \Phi(X)$, $z \in \mathbb{R}^{N_z}$
where the optimal latent dimension was found to be $N_z = 32$ which will be discussed in result section. The encoder applies successive one-dimensional convolutional layers with stride two, progressively reducing the spatial resolution while extracting hierarchical spatial features before projecting the representation into the 32-dimensional latent space. The decoder, denoted by $\Psi$, mirrors this structure and reconstructs the full-order flow field as $\hat{X} = \Psi(z)$, with the original dimension $(3,1000)$.

The autoencoder is trained by minimizing the mean squared reconstruction error over the training dataset. For a dataset containing $N$ snapshots, the reconstruction loss is defined as
\begin{equation}
\mathcal{L}_{AE} 
= \frac{1}{N} \sum_{i=1}^{N} 
\left\| X^{(i)} - \Psi\big(\Phi(X^{(i)})\big) \right\|_2^2,
\end{equation}
where $X^{(i)}$ denotes the $i$-th high-fidelity flow snapshot and $\|\cdot\|_2$ represents the Euclidean norm over all spatial locations and physical variables. Minimizing this loss enables the network to learn a nonlinear reduced manifold that preserves essential flow structures such as shock fronts, contact discontinuities, and rarefaction waves while significantly reducing dimensionality.

Although the autoencoder effectively compresses and reconstructs individual flow states, it does not model the mapping between the initial and final states of the system. To capture this evolution, a forward operator is introduced in the latent space, as illustrated in Figure~\ref{fig:ae_rom_workflow}. Let $X_0$ denote the initial flow state and $X_f$ denote the final flow state at time $t=0.2$. Their corresponding latent representations are defined as
$z_0 = \Phi(X_0)$, $z_T = \Phi(X_f)$.
After training the autoencoder, the encoder $\Phi$ is frozen to ensure that the latent manifold remains fixed. The forward operator $F$ is then trained to approximate the mapping
$ z_T \approx F(z_0)$ thus learning the evolution of the system directly within the reduced space. The latent forward operator is trained using paired initial and final latent states. 
For the $i$-th training sample, the encoded initial and final states are given by
$z_0^{(i)}=\Phi(X_0^{(i)})$ and $z_f^{(i)}=\Phi(X_f^{(i)})$, respectively. 
The forward-operator loss is defined as
\begin{equation}
\mathcal{L}_{F}
=
\frac{1}{N}
\sum_{i=1}^{N}
\left\|
z_f^{(i)}
-
\mathcal{F}
\left(
z_0^{(i)}
\right)
\right\|_2^2 ,
\end{equation}
where $\mathcal{F}$ denotes the learned latent-space forward operator. 
Minimizing this loss enables the model to learn the reduced mapping from the initial latent state to the final-time latent state.

During prediction, the complete AE-ROM operates in three sequential stages, as shown in Figure~\ref{fig:ae_rom_workflow}. First, a new initial condition $X_0(\boldsymbol{\theta})$ is encoded to obtain its latent representation $z_0$. Second, the forward operator predicts the latent final state $z_T$. Third, the decoder reconstructs the full-order final flow field as $\hat{X}_f = \Psi(z_T)$. Because this procedure consists solely of forward evaluations of trained neural networks, it eliminates the need for repeated high-fidelity simulations. This computational efficiency is essential for embedding the surrogate model within the Bayesian inference framework, where a large number of forward evaluations are required for posterior sampling.

\subsection{Mathematical Formulation of the Bayesian Inverse Problem}

With the autoencoder-based reduced-order surrogate established, the inverse problem is formulated as a parameter-to-observation mapping. 
The objective is to infer the unknown initial density and pressure states of the Sod shock tube from sparse and noisy observations of the final-time flow field. 
The unknown parameter vector is defined as
\begin{equation}
\boldsymbol{\theta}
=
\left[
\rho_L,\;
p_L,\;
\rho_R,\;
p_R
\right]^T ,
\end{equation}
where $\rho_L$ and $p_L$ denote the left-state density and pressure, and $\rho_R$ and $p_R$ denote the right-state density and pressure. 
The initial velocity is fixed to zero on both sides of the diaphragm and is therefore not included in the inverse parameter vector. For a given parameter vector $\boldsymbol{\theta}$, the corresponding initial shock-tube state is constructed as
\begin{equation}
X_0(x;\boldsymbol{\theta})
=
\begin{cases}
\left(\rho_L,\;0,\;p_L\right), & x \leq 0.5, \\
\left(\rho_R,\;0,\;p_R\right), & x > 0.5 .
\end{cases}
\end{equation}
Thus, each candidate parameter vector uniquely defines a piecewise-constant initial condition. Let $\mathcal{G}$ denote the high-fidelity forward solver that advances the initial condition to the final-time flow field,
\begin{equation}
X_f(\boldsymbol{\theta})
=
\mathcal{G}
\left(
X_0(\boldsymbol{\theta})
\right).
\end{equation}
Directly evaluating $\mathcal{G}$ inside an MCMC sampler is computationally expensive because posterior sampling requires a large number of repeated forward evaluations. 
Therefore, the high-fidelity solver is replaced by the trained autoencoder-based reduced-order surrogate. The reduced-order surrogate first maps the initial condition to a latent representation using the encoder $\Phi$,
\begin{equation}
z_0
=
\Phi
\left(
X_0(\boldsymbol{\theta})
\right).
\end{equation}
The latent forward operator $\mathcal{F}$ then predicts the final-time latent state,
\begin{equation}
\hat{z}_f
=
\mathcal{F}
\left(
z_0
\right),
\end{equation}
and the decoder $\Psi$ reconstructs the predicted final-time flow field,
\begin{equation}
\hat{X}_f(\boldsymbol{\theta})
=
\Psi
\left(
\hat{z}_f
\right).
\end{equation}
Combining these operations, the complete surrogate mapping is written as
\begin{equation}
\hat{X}_f(\boldsymbol{\theta})
=
\Psi
\left[
\mathcal{F}
\left(
\Phi
\left(
X_0(\boldsymbol{\theta})
\right)
\right)
\right].
\end{equation}
This expression defines the parameter-to-field mapping used inside the Bayesian inverse problem. 
It maps a proposed set of initial density and pressure states directly to a predicted final-time flow field without requiring a high-fidelity CFD solve.

\begin{figure*}[ht!]
    \centering
    \includegraphics[width=0.75\textwidth]{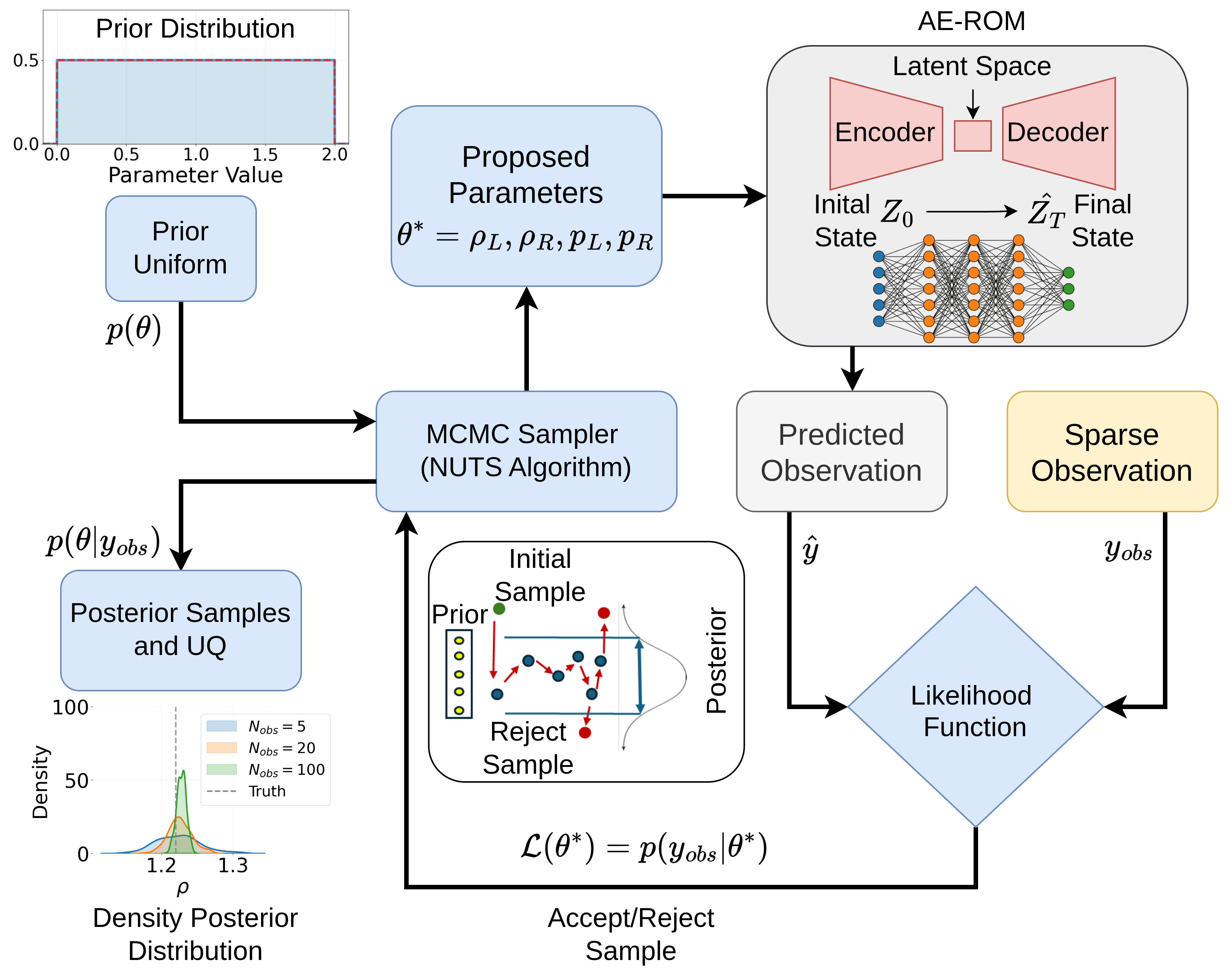}
    \caption{Schematic of the AE-ROM-enabled Bayesian inversion framework. Candidate initial-state parameters are drawn from uniform priors and passed through the AE-ROM surrogate to generate predicted observations. These predictions are compared with sparse noisy measurements through the likelihood function. The NUTS sampler uses this likelihood to explore the posterior distribution, producing posterior samples and uncertainty estimates for the inferred initial density and pressure states.}
    \label{fig:bayesian_framework}
\end{figure*}

Since the available measurements are sparse, the full predicted final-time field is not compared at every grid point. 
Instead, an observation operator $\mathcal{H}$ is introduced to extract only the measured quantities from the predicted field. 
In this work, the observations consist of density and pressure at selected spatial locations. 
The model-predicted observation vector is therefore
\begin{equation}
\mathcal{M}(\boldsymbol{\theta})
=
\mathcal{H}
\left[
\hat{X}_f(\boldsymbol{\theta})
\right].
\end{equation}
Here, $\mathcal{M}(\boldsymbol{\theta})$ contains the AE-ROM-predicted density and pressure values at the observation points. 
If observations are collected at $N_{\mathrm{obs}}$ spatial locations and both density and pressure are measured, then the total dimension of the observation vector is
\begin{equation}
N_y = 2N_{\mathrm{obs}} .
\end{equation}
The synthetic observation vector is generated by adding measurement noise to the high-fidelity final-time solution,
\begin{equation}
\mathbf{y}_{obs}
=
\mathcal{H}
\left[
X_f(\boldsymbol{\theta}_{true})
\right]
+
\boldsymbol{\epsilon},
\end{equation}
where $\boldsymbol{\theta}_{true}$ denotes the ground-truth parameter vector and $\boldsymbol{\epsilon}$ represents measurement noise. 
The noise is assumed to be independent and Gaussian,
\begin{equation}
\boldsymbol{\epsilon}
\sim
\mathcal{N}
\left(
\mathbf{0},
\sigma^2 \mathbf{I}
\right),
\end{equation}
where $\sigma$ is the prescribed noise standard deviation. Using this Gaussian noise model, the likelihood function is defined as
\begin{equation}
p
\left(
\mathbf{y}_{obs}
\mid
\boldsymbol{\theta}
\right)
=
\frac{1}
{
\left(2\pi\sigma^2\right)^{N_y/2}
}
\exp
\left[
-
\frac{1}{2\sigma^2}
\left\|
\mathbf{y}_{obs}
-
\mathcal{M}
\left(
\boldsymbol{\theta}
\right)
\right\|_2^2
\right].
\end{equation}
This likelihood assigns higher probability to parameter values whose surrogate-predicted observations are close to the sparse noisy measurements. For numerical stability, the logarithm of the likelihood is evaluated during sampling. 
Ignoring constants independent of $\boldsymbol{\theta}$, the log-likelihood is written as
\begin{equation}
\log
p
\left(
\mathbf{y}_{obs}
\mid
\boldsymbol{\theta}
\right)
\propto
-
\frac{1}{2\sigma^2}
\sum_{i=1}^{N_y}
\left[
y_{obs,i}
-
\mathcal{M}_i
\left(
\boldsymbol{\theta}
\right)
\right]^2 .
\end{equation}
This expression shows that the Bayesian inverse problem penalizes the squared mismatch between the observed data and the AE-ROM-predicted measurements. The prior distribution encodes the admissible ranges of the unknown initial states. 
Independent uniform priors are assigned to each parameter:
\begin{equation}
\rho_L,\;p_L
\sim
\mathcal{U}(0,2),
\qquad
\rho_R,\;p_R
\sim
\mathcal{U}(0,0.2).
\end{equation}
These prior ranges contain the parameter space used to generate the high-fidelity training and testing data while allowing a broader range of physically admissible states. The posterior distribution is obtained using Bayes' theorem,
\begin{equation}
p
\left(
\boldsymbol{\theta}
\mid
\mathbf{y}_{obs}
\right)
=
\frac{
p
\left(
\mathbf{y}_{obs}
\mid
\boldsymbol{\theta}
\right)
p
\left(
\boldsymbol{\theta}
\right)
}
{
p
\left(
\mathbf{y}_{obs}
\right)
}
\propto
p
\left(
\mathbf{y}_{obs}
\mid
\boldsymbol{\theta}
\right)
p
\left(
\boldsymbol{\theta}
\right),
\end{equation}
where $p(\boldsymbol{\theta} \mid \mathbf{y}_{obs})$ is the posterior distribution, $p(\mathbf{y}_{obs} \mid \boldsymbol{\theta})$ is the likelihood, $p(\boldsymbol{\theta})$ is the prior, and $p(\mathbf{y}_{obs})$ is the evidence. 
The evidence acts as a normalization constant and is not required for MCMC sampling. Because the posterior distribution is nonlinear and cannot be evaluated analytically, samples are drawn using the No-U-Turn Sampler (NUTS), an adaptive Hamiltonian Monte Carlo method. 
At each sampling step, a candidate parameter vector $\boldsymbol{\theta}^{*}$ is proposed, the corresponding initial condition $X_0(\boldsymbol{\theta}^{*})$ is constructed, and the AE-ROM surrogate is used to generate the predicted observations $\mathcal{M}(\boldsymbol{\theta}^{*})$. 
The likelihood of the proposed sample is then evaluated by comparing these predicted observations with the sparse noisy data. 
Replacing the high-fidelity solver with the trained surrogate makes repeated posterior sampling computationally tractable. The complete Bayesian inversion workflow is illustrated in Fig.~\ref{fig:bayesian_framework}. 
Candidate initial-state parameters are drawn from the prior distribution, propagated through the AE-ROM surrogate, compared with sparse noisy observations through the likelihood function, and accepted or rejected by the NUTS sampler to generate posterior samples. After sampling, the posterior distribution is represented by a collection of samples,
\begin{equation}
\left\{
\boldsymbol{\theta}^{(s)}
\right\}_{s=1}^{N_s},
\end{equation}
where $N_s$ is the number of posterior samples. 
Each sample corresponds to one possible set of initial density and pressure states. 
For each posterior sample, the corresponding initial density and pressure fields are reconstructed from the piecewise initial-condition definition. For a generic inferred field $q(x)$, where $q$ denotes either density or pressure, the posterior mean is computed pointwise as
\begin{equation}
\bar{q}(x_j)
=
\frac{1}{N_s}
\sum_{s=1}^{N_s}
q^{(s)}(x_j),
\end{equation}
where $x_j$ denotes the $j$-th spatial grid point. 
The posterior standard deviation is computed as
\begin{equation}
\sigma_q(x_j)
=
\left[
\frac{1}{N_s-1}
\sum_{s=1}^{N_s}
\left(
q^{(s)}(x_j)
-
\bar{q}(x_j)
\right)^2
\right]^{1/2}.
\end{equation}
The 95\% credible interval is obtained pointwise from the 2.5th and 97.5th percentiles of the posterior samples. The posterior mean error is quantified using the root-mean-square error,
\begin{equation}
\mathrm{RMSE}_q
=
\left[
\frac{1}{N_x}
\sum_{j=1}^{N_x}
\left(
\bar{q}(x_j)
-
q_{true}(x_j)
\right)^2
\right]^{1/2},
\end{equation}
where $N_x$ is the number of spatial grid points and $q_{true}$ is the ground-truth initial field. 
The mean posterior uncertainty is defined as the spatial average of the posterior standard deviation,
\begin{equation}
\overline{\sigma}_q
=
\frac{1}{N_x}
\sum_{j=1}^{N_x}
\sigma_q(x_j).
\end{equation}
The RMSE measures the accuracy of the posterior mean estimate, while $\overline{\sigma}_q$ measures the average uncertainty remaining in the inferred initial field. 
These metrics are used to evaluate how observation density affects posterior accuracy and uncertainty contraction. It should be noted that the posterior uncertainty reported in this study is conditioned on the trained AE-ROM surrogate and the assumed Gaussian observation model. 
An additional surrogate-model discrepancy term is not included in the likelihood. 
Therefore, the reported credible intervals quantify uncertainty under the surrogate model and prescribed observation noise, while explicit treatment of surrogate-model error is left for future work.

\section{Results and Discussion}
In this section, we present a comprehensive evaluation of the proposed framework. We begin by assessing the accuracy of the AE-ROM as a predictive surrogate, including the impact of latent space dimensionality and snapshot number. We then demonstrate its application within the Bayesian framework to solve the inverse problem of recovering unknown initial conditions from sparse data. Finally, we conduct a parametric study on the effect of observation size to analyze the inverse framework performance.

\subsection{AE-ROM Performance and Fidelity}

The first stage of the proposed framework constructs a non-intrusive reduced-order representation of high-dimensional compressible flow fields using a convolutional autoencoder. For hyperbolic systems such as the Euler equations, the reduced manifold must simultaneously (i) preserve sharp discontinuities and (ii) maintain strong compression to enable efficient downstream Bayesian inference. This section evaluates the reconstruction accuracy, latent dimension sensitivity, and data efficiency of the AE-ROM.

\subsubsection{Latent Dimension Trade-off and Convergence}

To identify an appropriate bottleneck size for the reduced-order manifold, we performed a parametric sweep over the latent dimension, $N_z \in \{4, 8, 16, 32, 64\}$. Model performance was evaluated using the validation mean-squared error (MSE) computed across the density $\rho$, velocity $u$, and pressure $p$ fields.

\begin{figure}[h]
    \centering
    \includegraphics[width=0.95\textwidth]{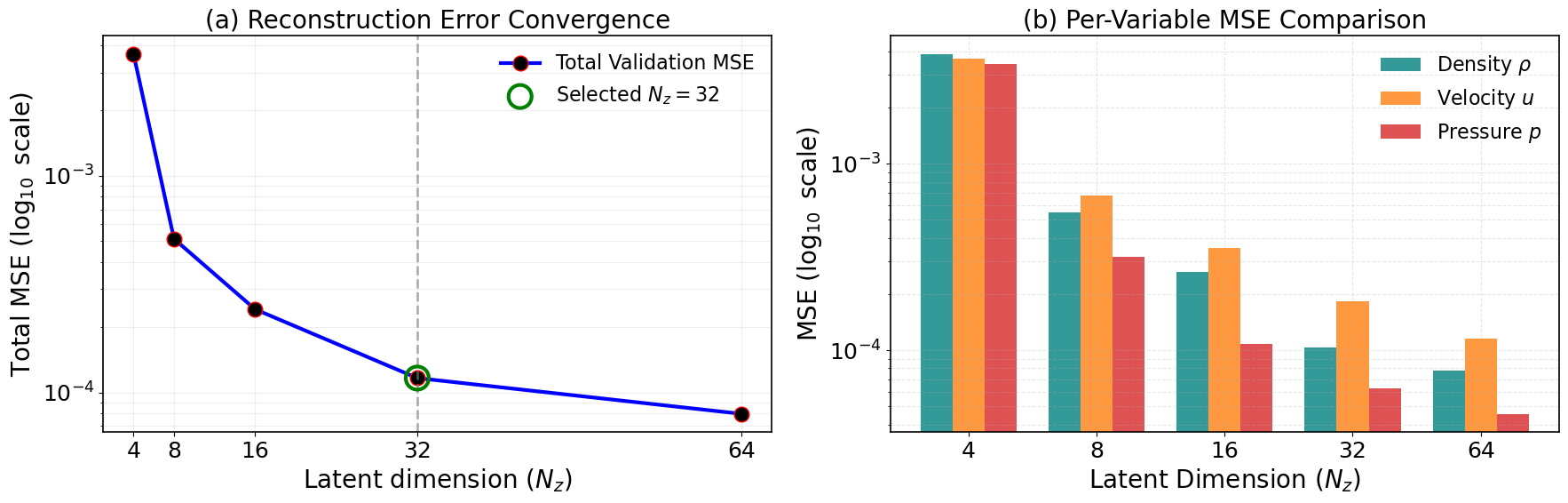}
    \caption{Effect of latent dimension on AE-ROM reconstruction accuracy. 
(a) Total validation MSE decreases rapidly with increasing $N_z$ and exhibits an elbow near the selected value of $N_z=32$. 
(b) Per-variable validation MSE for density $\rho$, velocity $u$, and pressure $p$, showing consistent reconstruction improvement across all flow variables.}
    \label{fig:mse_convergence}
\end{figure}

As shown in Fig.~\ref{fig:mse_convergence}(a), the total validation MSE decreases rapidly as the latent dimension increases. The largest improvement occurs when the latent dimension is increased from $N_z=4$ to $N_z=8$, corresponding to an error reduction of $86.01\%$. This substantial decrease indicates that very low-dimensional latent spaces are insufficient to represent the dominant wave structures of the Riemann problem. Further increases to $N_z=16$ and $N_z=32$ continue to improve reconstruction accuracy, with relative error reductions of $52.70\%$ and $51.87\%$, respectively. However, increasing the latent dimension from $N_z=32$ to $N_z=64$ produces a smaller relative improvement of $31.62\%$, indicating diminishing returns.

The per-variable MSE comparison in Fig.~\ref{fig:mse_convergence}(b) shows a consistent decrease in reconstruction error for all three flow variables as the latent dimension increases. This confirms that the reduction in total MSE is not driven by a single variable, but reflects improved reconstruction of density, velocity, and pressure simultaneously. At larger latent dimensions, pressure exhibits the lowest reconstruction error, while velocity remains slightly more difficult to reconstruct, likely due to sharper localized variations near the wave interaction regions.

\begin{figure}[h]
    \centering
    \includegraphics[width=0.99\textwidth]{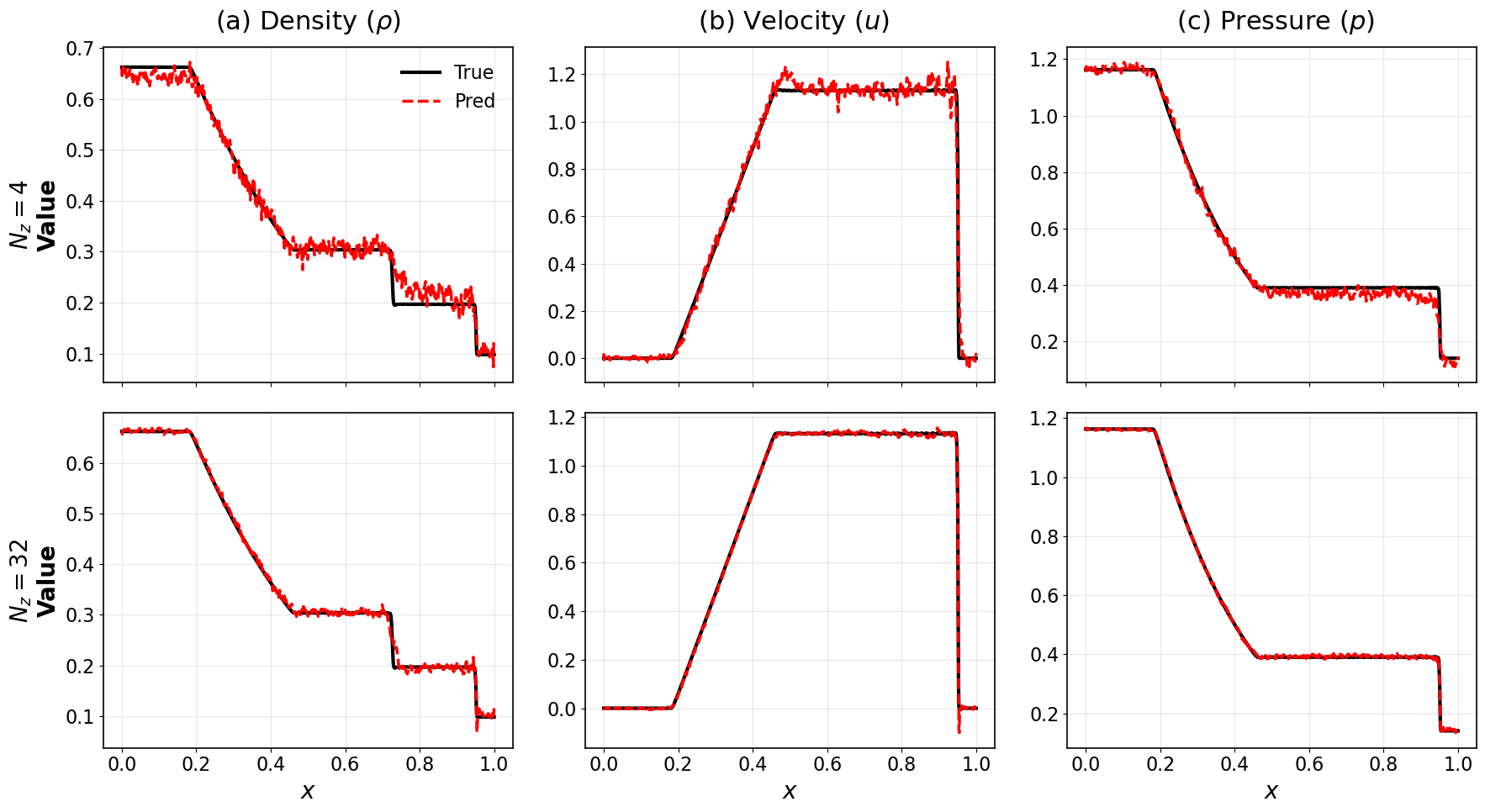}
    \caption{Qualitative comparison of AE-ROM reconstructions for latent dimensions 4 and 32. (Top Row) $N_z=4$ shows that the highly compressed latent space captures the overall flow structure but produces visible oscillations and smeared discontinuities. (Bottom Row) $N_z=32$ demonstrates that the selected latent dimension of 32 closely matches the high-fidelity density, velocity, and pressure fields while better preserving the rarefaction wave, contact discontinuity, and shock structure.}
    \label{fig:recon_comparison}
\end{figure}

The qualitative effect of latent dimension is shown in Fig.~\ref{fig:recon_comparison}, where reconstructions obtained with $N_z=4$ and $N_z=32$ are compared against the high-fidelity solution for density, velocity, and pressure. For $N_z=4$, the autoencoder captures the overall shape of the solution but exhibits visible oscillations, smeared discontinuities, and reduced accuracy near the shock and contact discontinuity. These errors are most apparent in the density field, where the intermediate plateau and sharp jumps are not fully resolved. In contrast, the $N_z=32$ reconstruction closely follows the high-fidelity solution across all three variables. The shock location, rarefaction profile, contact discontinuity, and post-shock states are recovered with substantially improved accuracy. This comparison confirms that the selected latent dimension provides sufficient representational capacity to preserve the key wave structures of the shock-tube solution.

Based on the trade-off between reconstruction accuracy and latent-space complexity, $N_z=32$ is selected as the latent dimension for the remainder of this study. This choice provides high reconstruction fidelity while maintaining a compact reduced representation suitable for efficient latent-space forward modeling and Bayesian inference.

\subsubsection{Scaling Study: Effect of Training Data Volume}
To evaluate the data efficiency of the proposed AE-ROM, we conducted a scaling study by varying the number of high-fidelity simulations used for training while keeping the latent dimension fixed at $N_z = 32$. The training set size was varied from 20 to 500 ($n_{\mathrm{sim}} \in \{20, 50, 100, 150, 200, 250, 300, 350, 400, 450, 500\}$) simulations. The objective was to determine the minimum amount of high-fidelity CFD data required to achieve accurate manifold reconstruction without adding unnecessary training cost.

\begin{figure}[ht!]
    \centering
    \includegraphics[width=0.65\textwidth]{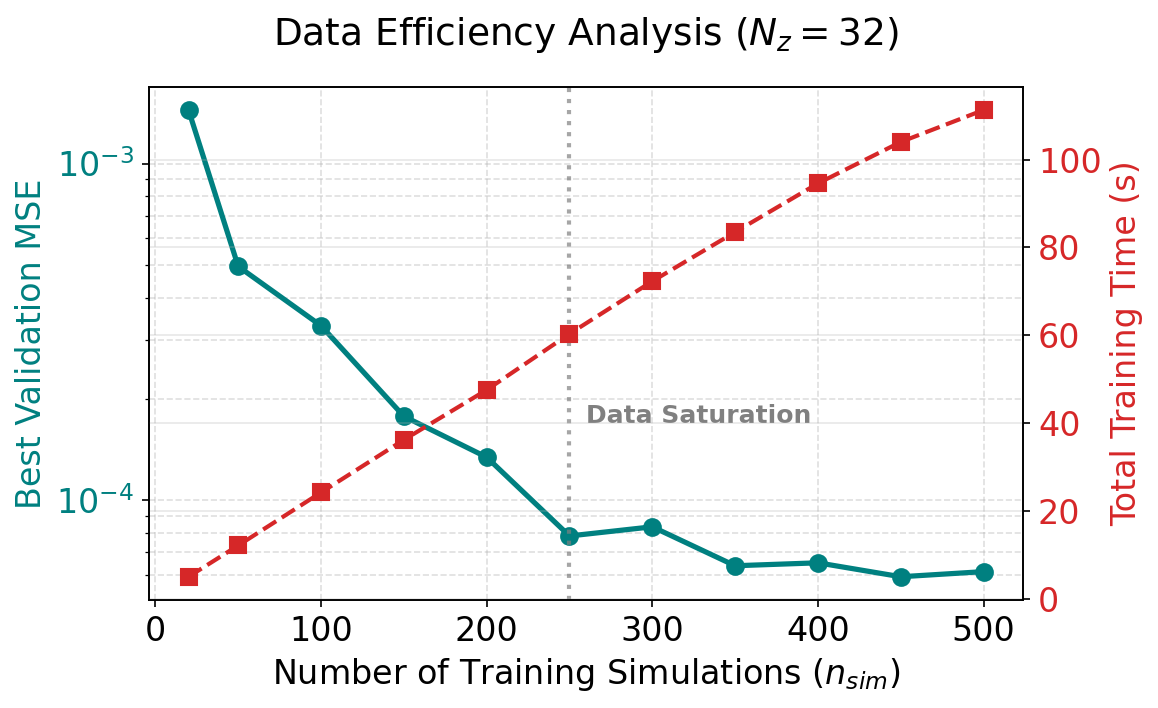}
    \caption{Data efficiency analysis for the AE-ROM with fixed latent dimension of $N_z=32$. The validation MSE (left axis) shows a rapid decay up to $n_{sim}=250$, after which it reaches a saturation point of diminishing returns. The total training time (right axis) increases nearly linearly with the training data size. The vertical dotted line marks the optimal data budget of 250 simulations, representing a practical balance between reconstruction accuracy and computational cost.}
    \label{fig:scaling_study}
\end{figure}

As shown in Fig.~\ref{fig:scaling_study}, the validation MSE decreases rapidly as the number of training simulations increases from 20 to 250. Over this range, the model benefits significantly from additional training data, with the reconstruction error approaching the order of $\mathcal{O}(10^{-4})$. Beyond $n_{\mathrm{sim}}=250$, however, the validation error begins to saturate, and additional simulations provide only marginal improvements in reconstruction accuracy. In contrast, the total training time continues to increase nearly linearly with the size of the training dataset.

This behavior indicates a clear trade-off between reconstruction accuracy and computational cost. Although larger datasets continue to slightly reduce the validation error, the improvement beyond 250 simulations is relatively small compared to the added training expense. Therefore, $n_{\mathrm{sim}}=250$ is selected as the data budget for the subsequent AE-ROM and Bayesian inversion experiments. This result demonstrates that the proposed framework can achieve high reconstruction fidelity using a relatively modest number of high-fidelity simulations.

\subsubsection{Predictive Latent Evolution (Forward Operator)}

After selecting the latent dimension $N_z=32$ and the training data budget $n_{\mathrm{sim}}=250$, the final step in constructing the AE-ROM is to train the latent-space forward operator. The purpose of this operator is to learn the mapping from the encoded initial state to the encoded final state, $\mathcal{F}: z_0 \mapsto z_T$, where $z_0=\Phi(X_0)$ is the latent representation of the initial condition and $z_T=\Phi(X_f)$ is the latent representation of the final-time solution at $t=0.2$ s. Once trained, this forward operator enables the AE-ROM to predict the full final-time flow field directly from a new initial condition without requiring a high-fidelity CFD simulation.

\begin{figure}[htbp]
    \centering
    \includegraphics[width=0.65\textwidth]{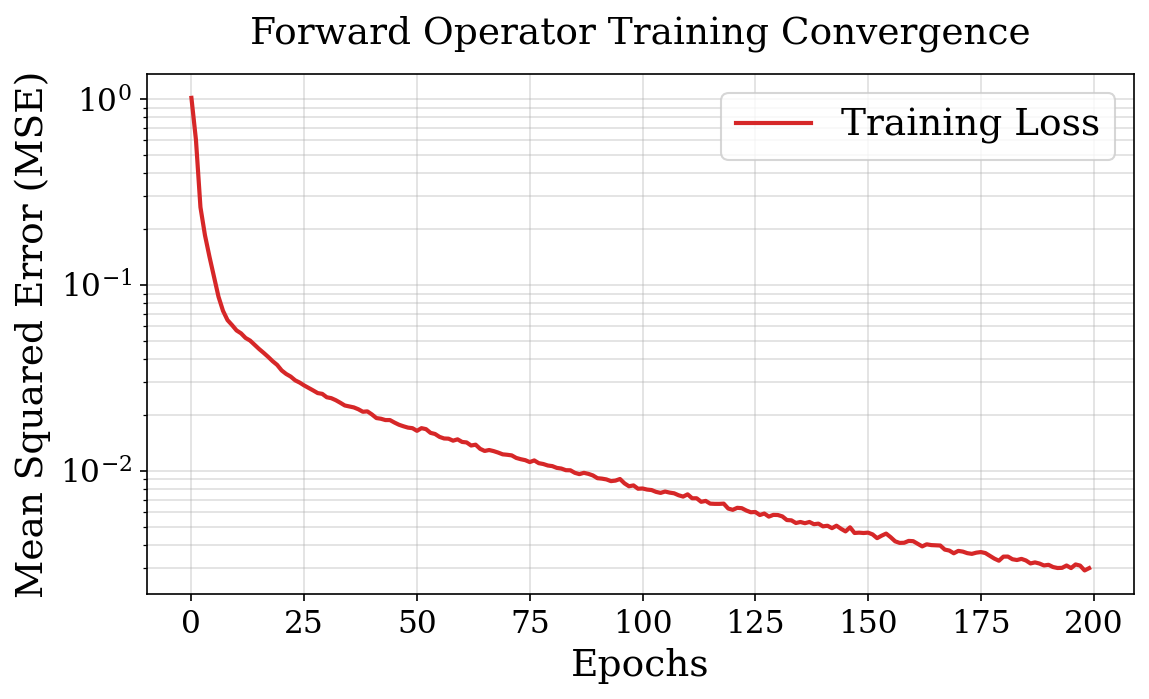}
    \caption{Training convergence of the latent-space forward operator. The MSE decreases steadily over 200 epochs, indicating stable learning of the mapping from the initial latent state $z_0$ to the final latent state $z_T$ using the selected data budget of $n_{\mathrm{sim}}=250$.}
    \label{fig:forward_loss}
\end{figure}

The forward operator was trained for 200 epochs using the encoded initial and final states from the selected 250-simulation dataset. As shown in Fig.~\ref{fig:forward_loss}, the training loss decreases steadily over the full training window, indicating stable convergence of the latent-space mapping. The rapid initial decrease followed by a gradual reduction in MSE suggests that the operator first learns the dominant input-output relationship and then progressively refines the latent prediction. This behavior indicates that the autoencoder provides a structured reduced space in which the final-time evolution can be learned efficiently.

\begin{figure*}[ht!]
    \centering
    \includegraphics[width=0.85\textwidth]{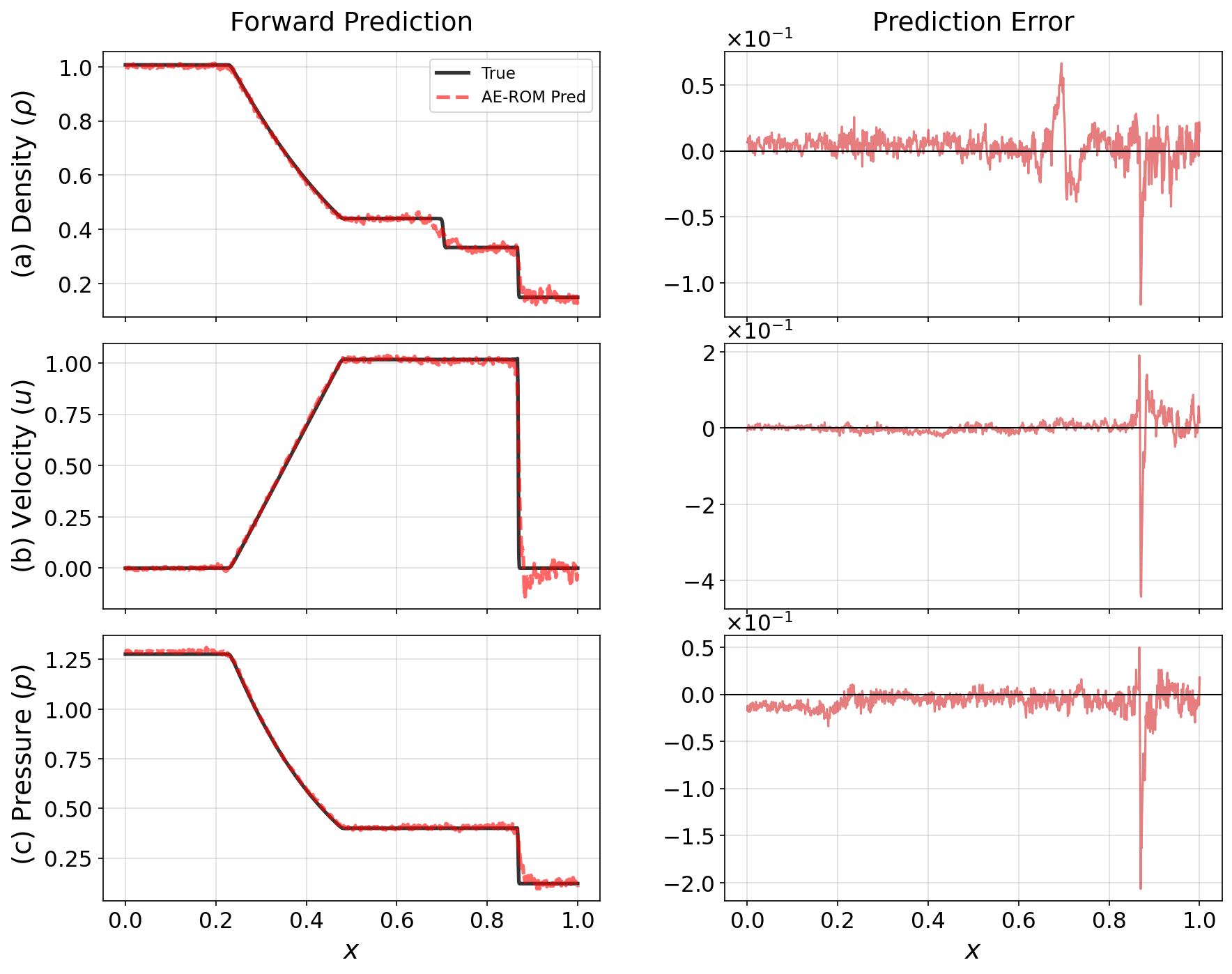}
     \caption{Forward prediction performance of the AE-ROM on a held-out test case at $t=0.2$ s. The left column compares the AE-ROM prediction with the high-fidelity CFD solution for density, velocity, and pressure. The right column shows the corresponding spatial prediction error. The surrogate accurately captures the main wave structures, including the rarefaction region, contact discontinuity, and shock location, while the largest errors remain localized near discontinuities.}
    \label{fig:forward_prediction}
\end{figure*}

To evaluate the predictive capability of the complete AE-ROM, the trained forward operator was tested on a held-out initial condition not used during training. The predicted latent final state was decoded to obtain the reconstructed density, velocity, and pressure fields at $t=0.2$ s. Figure~\ref{fig:forward_prediction} compares the AE-ROM prediction with the high-fidelity CFD solution and also reports the corresponding spatial prediction error.

The left column of Fig.~\ref{fig:forward_prediction} shows that the AE-ROM prediction closely follows the high-fidelity solution for all three variables. The rarefaction profile, intermediate plateau regions, contact discontinuity, and shock location are all captured with good accuracy. The agreement is particularly strong in the smooth regions of the flow, where the predicted and true profiles nearly overlap. This demonstrates that the latent forward operator is able to propagate the dominant flow features from the initial condition to the final-time state.

The right column of Fig.~\ref{fig:forward_prediction} shows the prediction error ($x_{true} - \hat{x}_{pred}$) between the high-fidelity solution and the AE-ROM prediction. The errors remain small across most of the spatial domain and are primarily concentrated near discontinuities, especially around the shock and contact regions. This localization is expected for reduced-order approximations of hyperbolic problems, where small errors in discontinuity location can produce sharp residual peaks. Despite these localized errors, the overall prediction remains accurate, indicating that the AE-ROM provides a reliable surrogate for downstream Bayesian inversion.

\subsection{Bayesian Initial-State Inversion and Uncertainty Quantification}

In this section, we evaluate the performance of the AE-ROM surrogate within a Bayesian inversion framework. The objective is to infer the unknown initial density and pressure states on the left and right sides of the diaphragm from sparse and noisy observations of the final-time flow field. Specifically, the inferred parameters are the left and right density states, $\rho_L$ and $\rho_R$, and the left and right pressure states, $p_L$ and $p_R$. We first describe the observation configurations used in the inversion experiments, then examine how the posterior distributions evolve with observation density, and finally assess posterior predictive accuracy and uncertainty contraction.

\subsubsection{Observation Configurations and Noise Model}

\begin{figure*}[ht!]
    \centering
    \includegraphics[width=0.95\textwidth]{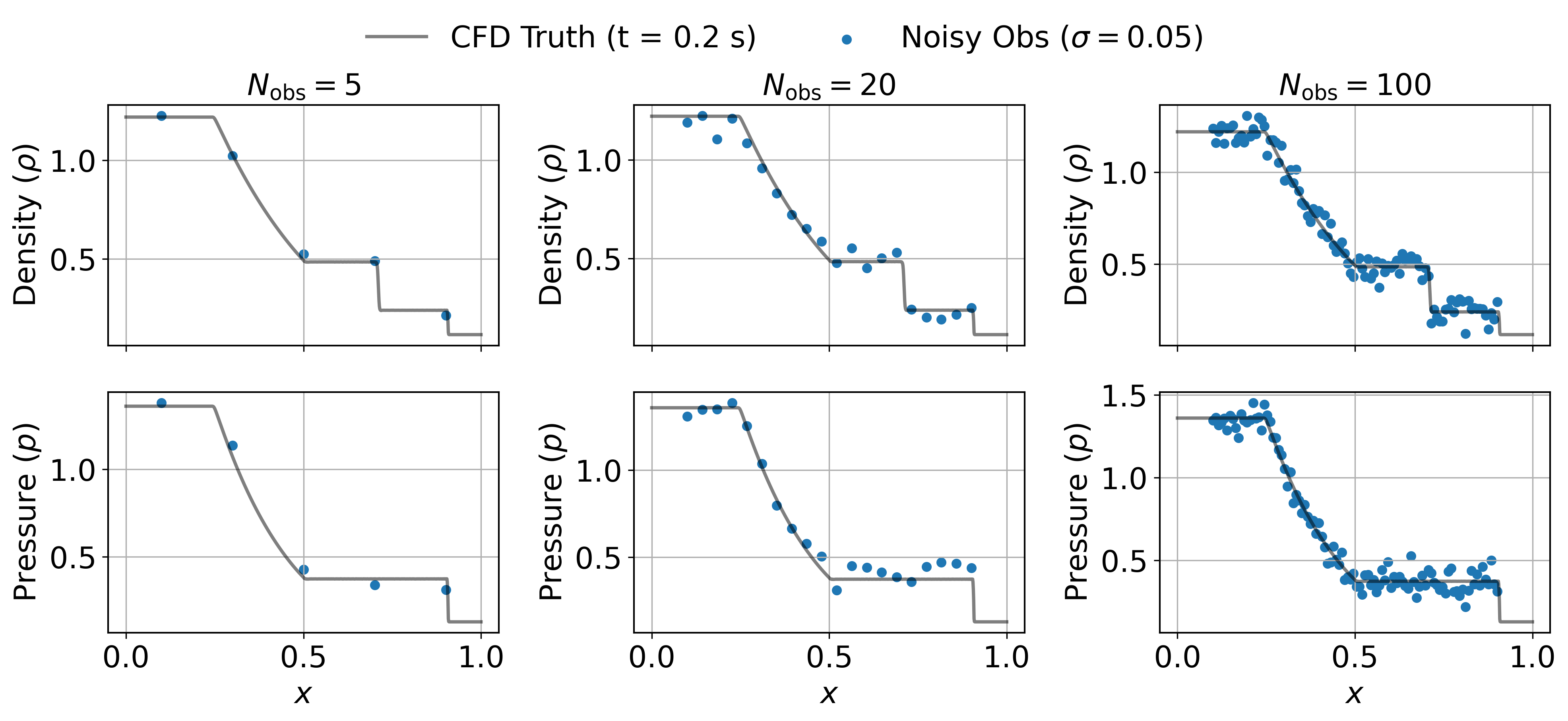}
    \caption{Synthetic observation configurations used for Bayesian inversion. 
    The first row shows density measurements and the second row shows pressure measurements 
    at $t = 0.2$ s for $N_{\mathrm{obs}} = 5$, 20, and 100. 
    Observation locations are uniformly distributed over $x \in [0.1,0.9]$ and 
    corrupted with independent Gaussian noise ($\sigma = 0.05$). 
    Increasing observation density progressively improves resolution of the rarefaction, 
    contact discontinuity, and shock structures.}
    \label{fig:observations}
\end{figure*}

Figure~\ref{fig:observations} shows the synthetic observation data used for Bayesian inversion. The first row corresponds to density observations and the second row corresponds to pressure observations at the final time, $t=0.2$ s. Three observation densities are considered: $N_{\mathrm{obs}}=5$, 20, and 100.

The observation points are uniformly distributed within the interior of the spatial domain, $x \in [0.1,0.9]$. The gray curves represent the high-fidelity CFD solution, while the blue markers denote noisy synthetic observations. Independent Gaussian noise with standard deviation $\sigma=0.05$ is added to the sampled measurements, consistent with the likelihood model used in the Bayesian formulation.

For the sparsest case, $N_{\mathrm{obs}}=5$, the observations provide only limited information about the final-time flow structure. In this setting, the rarefaction region, contact discontinuity, and shock location are only weakly constrained. Increasing the number of observations to $N_{\mathrm{obs}}=20$ provides a more detailed sampling of the solution profile, improving the information available to the inverse problem. For $N_{\mathrm{obs}}=100$, the observations resolve the main wave structures much more densely, thereby providing stronger constraints on the inferred initial states.

\subsubsection{Evolution of Posterior Distributions}

\begin{figure*}[ht!]
    \centering
    \includegraphics[width=0.95\textwidth]{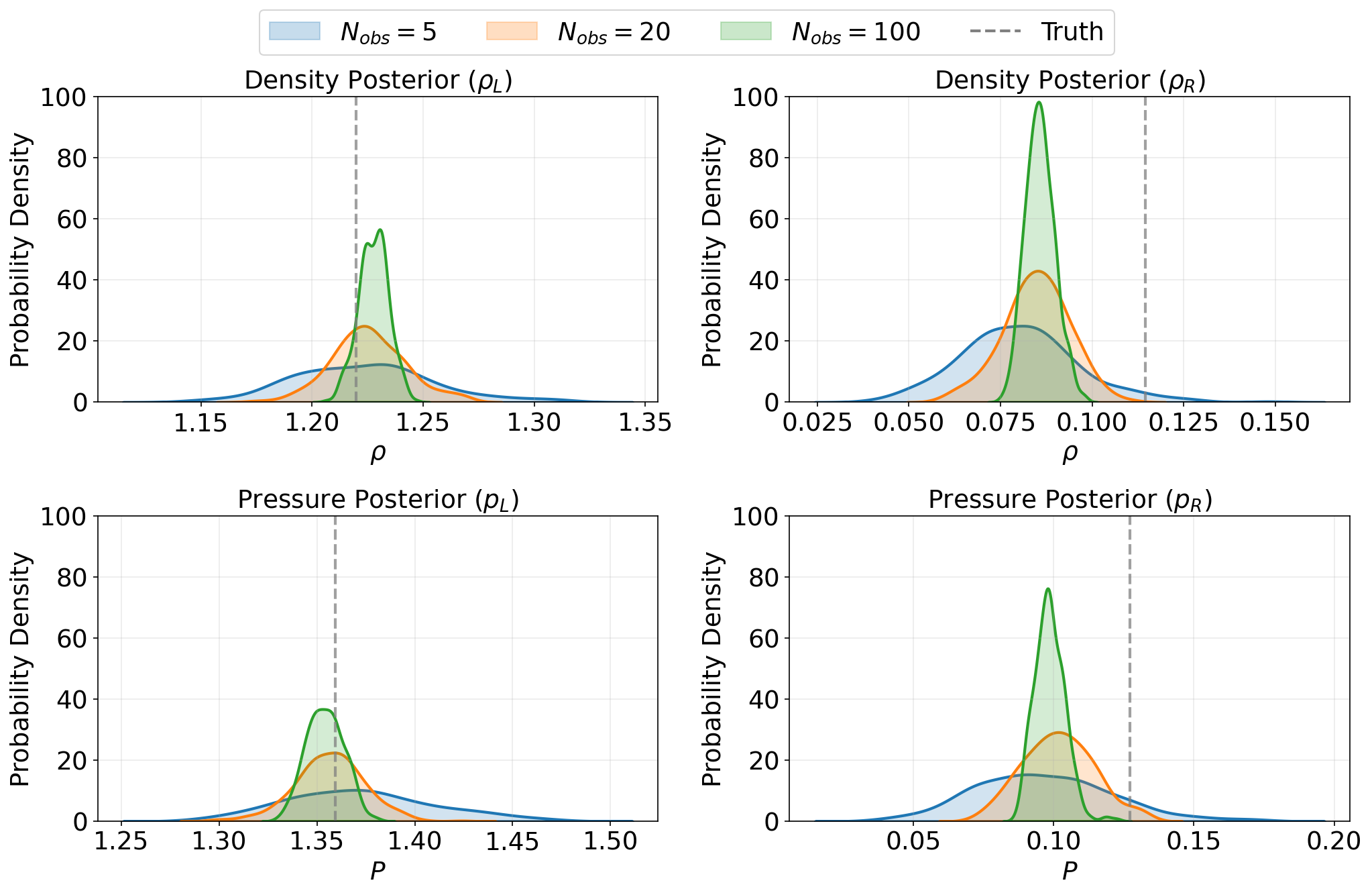}
    \caption{Posterior probability density functions of the inferred initial density and pressure parameters for increasing observation density. 
    The distributions are shown for $N_{\mathrm{obs}}=5$, 20, and 100, while the dashed vertical lines denote the corresponding ground-truth parameter values. 
    Increasing $N_{\mathrm{obs}}$ leads to progressive posterior contraction, indicating improved parameter identifiability and reduced uncertainty.}
    \label{fig:posterior_evolution}
\end{figure*}

Figure~\ref{fig:posterior_evolution} shows the posterior probability density functions of the inferred initial density and pressure parameters ($(\rho_L,\rho_R)$ and $(p_L,p_R)$) for the three observation densities. The dashed vertical lines indicate the corresponding ground-truth parameter values. For $N_{\mathrm{obs}}=5$, the posterior distributions are relatively broad, reflecting the limited information content of the sparse observations. Although the inferred distributions remain concentrated near physically plausible values, the uncertainty is large because only a small number of measurements are available to constrain the inverse problem.

As the number of observations increases to $N_{\mathrm{obs}}=20$, the posterior distributions become narrower, indicating improved parameter identifiability. The additional measurements provide more information about the final-time wave structure, allowing the Bayesian framework to reduce the range of admissible initial density and pressure states. For $N_{\mathrm{obs}}=100$, the posterior distributions contract substantially compared with the sparse-observation case. The density and pressure parameters become more sharply concentrated, showing that the inverse problem is more strongly constrained when the observation field is more densely sampled. This progressive narrowing of the posterior distributions demonstrates the direct influence of observation density on uncertainty reduction in the inferred initial states.

\subsubsection{Posterior Predictive Accuracy and Uncertainty Contraction}

\begin{figure*}[ht!]
\centering
\includegraphics[width=0.85\textwidth]{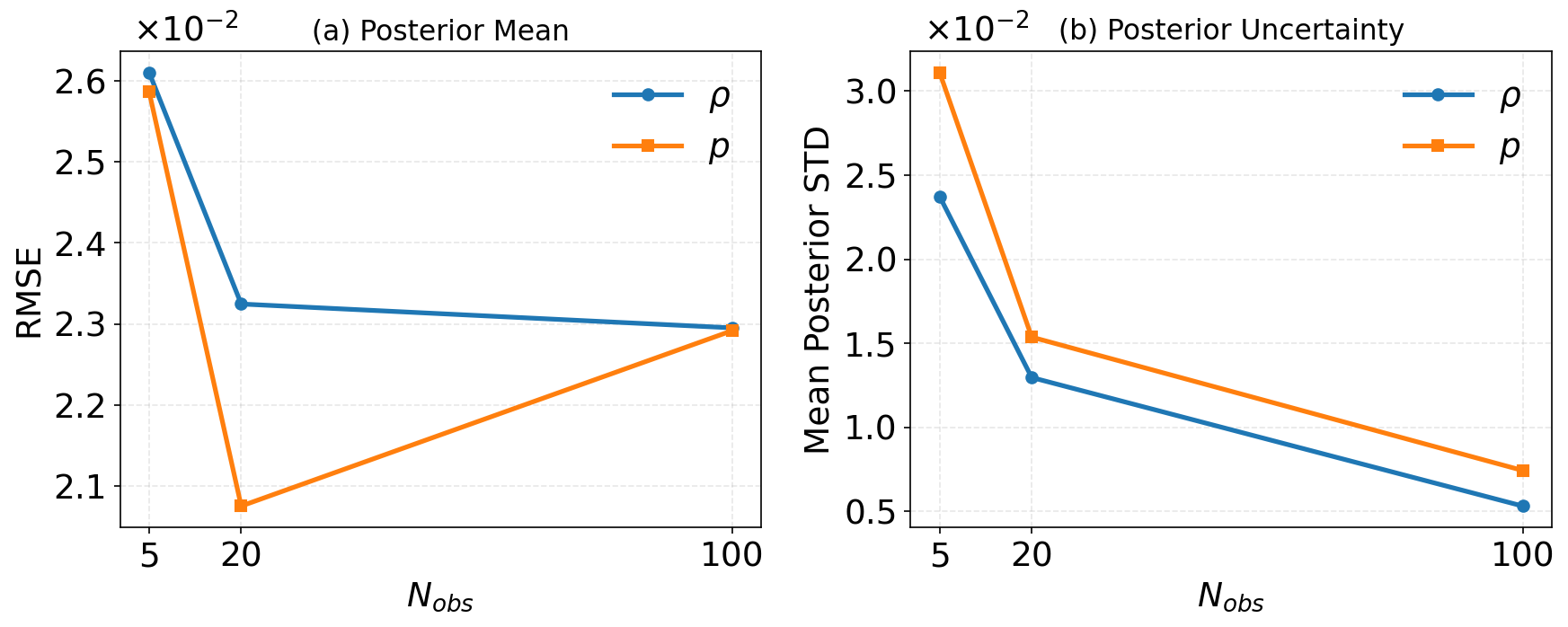}
\caption{
Effect of observation density on posterior predictive accuracy and uncertainty. 
(a) RMSE between the posterior mean and the ground-truth initial condition for density $\rho$ and pressure $p$. 
(b) Mean posterior standard deviation for density and pressure. 
Increasing $N_{\mathrm{obs}}$ produces a moderate reduction in posterior mean error but a substantial reduction in posterior uncertainty, indicating strong information gain from denser observations.
}
\label{fig:posterior_metrics}
\end{figure*}

Figure~\ref{fig:posterior_metrics} summarizes the effect of observation density on posterior predictive accuracy and uncertainty. The posterior mean error is measured using the root-mean-square error (RMSE), while posterior uncertainty is quantified using the mean posterior standard deviation.

As $N_{\mathrm{obs}}$ increases from 5 to 100, the RMSE decreases from $2.61\times10^{-2}$ to $2.30\times10^{-2}$ for density, corresponding to a reduction of approximately 12\%. For pressure, the RMSE decreases from $2.59\times10^{-2}$ to $2.29\times10^{-2}$, corresponding to a reduction of approximately 11\%. These results indicate that increasing the number of observations improves the posterior mean prediction, although the improvement in mean accuracy is moderate.

In contrast, the reduction in posterior uncertainty is much more pronounced. The mean posterior standard deviation decreases by 77.6\% for density and 76.1\% for pressure as the number of observations increases from 5 to 100. This strong reduction shows that the primary effect of increasing observation density is to improve confidence in the inferred state, rather than to produce a large shift in the posterior mean.

\begin{figure*}[ht!]
\centering
\includegraphics[width=0.95\textwidth]{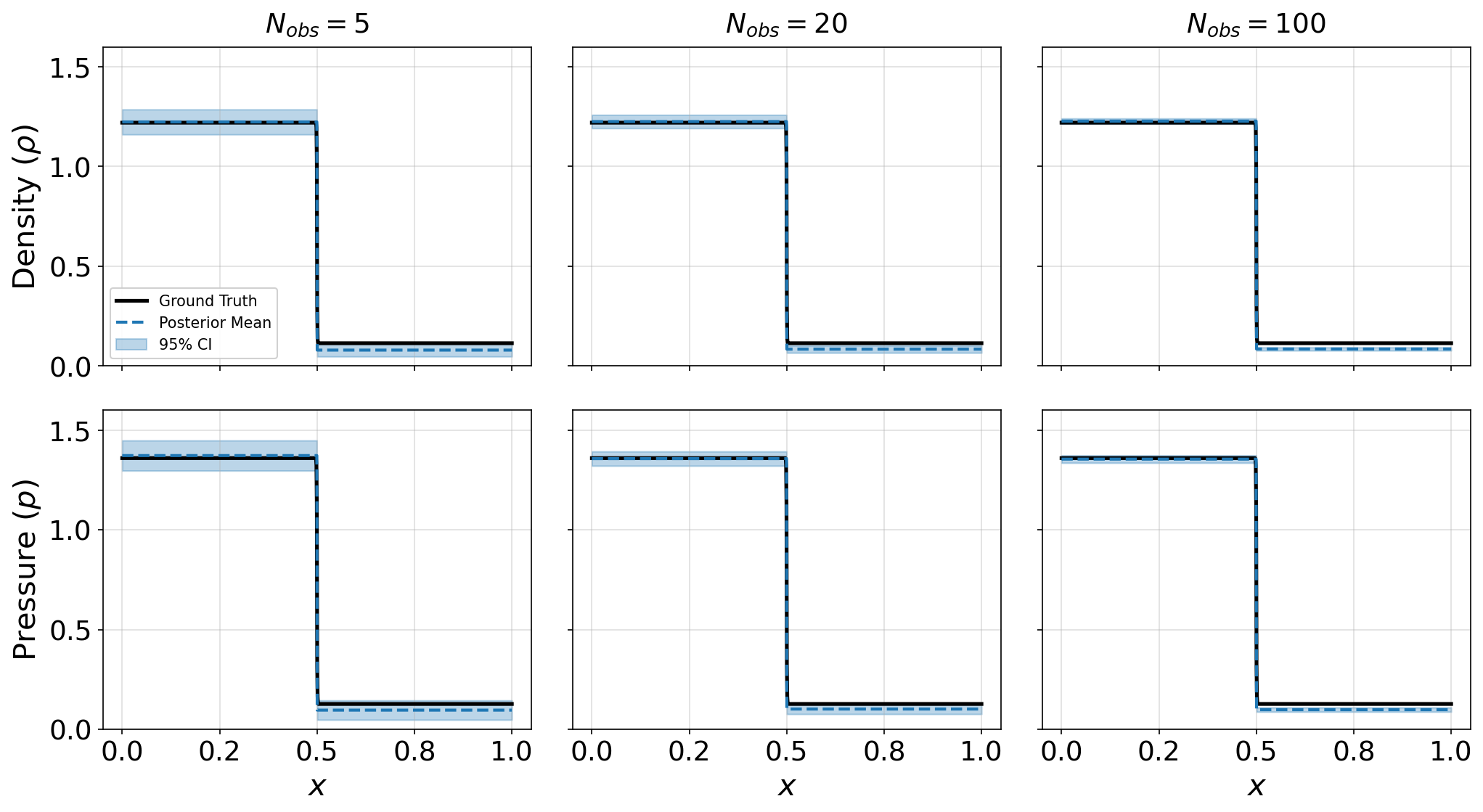}
\caption{
Spatial posterior predictive reconstructions of the inferred initial density (top row) and pressure (bottom row) fields for $N_{\mathrm{obs}}=5$, 20, and 100. 
The posterior mean is shown with dashed lines, the 95\% credible interval is shown by the shaded region, and the ground truth is shown with solid black lines. 
As the observation density increases, the credible intervals contract systematically while the posterior mean remains close to the ground-truth piecewise structure.
}
\label{fig:posterior_spatial}
\end{figure*}

The spatial posterior predictive reconstructions in Fig.~\ref{fig:posterior_spatial} further illustrate this behavior. For $N_{\mathrm{obs}}=5$, the 95\% credible intervals are relatively wide, reflecting the uncertainty associated with sparse measurements. As the observation density increases, the credible intervals contract systematically while the posterior mean remains close to the ground-truth profile. This behavior confirms that additional observations primarily reduce uncertainty and improve parameter identifiability, while preserving accurate recovery of the piecewise initial density and pressure structure.

It should be noted that the reported credible intervals are conditioned on the AE-ROM surrogate and the assumed Gaussian observation model. They do not explicitly include an additional surrogate-model discrepancy term, which will be considered in future extensions of the framework.

\section{Conclusions}

This paper developed a non-intrusive parametric reduced-order modeling framework for Bayesian inverse analysis of shock-dominated compressible flows under sparse and noisy observations. The proposed approach combines a convolutional autoencoder with a latent-space forward operator to construct an efficient AE-ROM surrogate. The autoencoder learns a compact nonlinear manifold for high-dimensional flow fields, while the forward operator maps encoded initial conditions to final-time latent states. This surrogate was then embedded within a NUTS-based Bayesian inference framework to recover unknown initial density and pressure states with quantified uncertainty.

The Sod shock tube problem was used as a controlled benchmark to demonstrate the complete workflow, including high-fidelity data generation, latent manifold construction, surrogate prediction, and posterior inference. A latent-dimension study showed that $N_z=32$ provides a practical balance between reconstruction accuracy and reduced-space compactness. A data-efficiency study further showed that accurate reconstructions can be achieved using a moderate training budget of 250 high-fidelity simulations. The trained latent forward operator accurately predicted final-time density, velocity, and pressure fields on held-out test cases, with the largest errors localized near discontinuities, as expected for hyperbolic flow problems.

The Bayesian inversion results demonstrated that the AE-ROM surrogate enables efficient posterior sampling for initial-state inference. Increasing the observation density from $N_{\mathrm{obs}}=5$ to $N_{\mathrm{obs}}=100$ produced substantial posterior uncertainty contraction while yielding moderate improvements in posterior mean accuracy. Specifically, the RMSE decreased by approximately 12\% for density and 11\% for pressure, while the mean posterior standard deviation decreased by 77.6\% for density and 76.1\% for pressure. These results indicate that additional observations primarily improve confidence and parameter identifiability rather than causing large shifts in the posterior mean. Overall, the proposed framework provides a computationally tractable and uncertainty-aware approach for inverse analysis in shock-dominated flow settings.

Several limitations remain. First, the present study is restricted to a one-dimensional Riemann problem with piecewise-constant initial conditions and zero initial velocity, which simplifies the inverse parameterization. Second, the latent forward operator is trained to predict a single final-time state, rather than learning time-continuous dynamics over arbitrary prediction horizons. Third, the Bayesian formulation assumes independent Gaussian observation noise and does not explicitly include surrogate-model discrepancy, which may lead to overconfident posteriors when surrogate errors are significant near discontinuities. Finally, although the autoencoder performs well for the present benchmark, latent representations may become more challenging to learn and stabilize in multidimensional flows with complex shock interactions and evolving flow topology.

Future work will extend the proposed framework in several directions. The first direction is the application to multidimensional compressible flows and reentry-vehicle configurations, where the inferred parameters may include aerodynamic stability coefficients and uncertain freestream or initial conditions. The second direction is the development of time-aware latent dynamics models, such as latent neural ODEs or neural operators, to improve temporal generalization and preserve important physical structures over longer horizons. The third direction is the incorporation of discrepancy-aware likelihood models and active learning strategies to improve posterior reliability while reducing the number of required high-fidelity simulations. These extensions will help advance the present proof-of-concept framework toward a robust Bayesian inverse-analysis tool for practical high-speed aerodynamics and digital-twin applications.
\section*{Acknowledgments}
The authors would like to thank the members of the Digital Twin Lab for their insightful feedback.

\section*{Code Availability}

The implementation of the proposed AE-ROM Bayesian inversion framework is publicly available on \href{https://github.com/bipintiwari2950/Bayesian_inversion_AE-ROM}{GitHub}. The repository includes scripts for Sod shock tube data generation, autoencoder and forward-operator training, Bayesian inference, and post-processing for reconstruction and uncertainty-quantification analysis.

\bibliographystyle{elsarticle-num}
\bibliography{references}

\end{document}